\documentclass[10pt,twocolumn,letterpaper]{article}

\usepackage{cvpr}
\usepackage{times}
\usepackage{epsfig}
\usepackage{graphicx}
\usepackage{amsmath}
\usepackage{amssymb}
\usepackage{caption}
\usepackage{subfig}
\usepackage{booktabs}
\usepackage{authblk}
\usepackage[pagebackref=true,breaklinks=true,letterpaper=true,colorlinks,bookmarks=false]{hyperref}

 \cvprfinalcopy % *** Uncomment this line for the final submission

\def\cvprPaperID{3981} % *** Enter the CVPR Paper ID here

\ifcvprfinal\fi

\def\0{{\bf 0}}
\def\1{{\bf 1}}

\def\MX{{\mathcal X}}

\def\MY{{\mathcal Y}}

\def\etal{{\em et al.\/}\,}

\begin{document}

\makeatletter
\renewcommand\AB@affilsepx{, \protect\Affilfont}
\makeatother
%%%%%%%%% TITLE
\title{Multi-shot Pedestrian Re-identification via Sequential Decision Making}

\author[1]{Jianfu Zhang}
\author[2]{Naiyan Wang}
\author[1]{Liqing Zhang}
\affil[1]{Shanghai Jiao Tong University\thanks{Representing Key Laboratory of Shanghai Education Commission for Intelligent Interaction and Cognitive Engineering, Department of Computer Science and Engineering, Shanghai Jiao Tong University.}}
\affil[2]{TuSimple \tt\small c.sis@sjtu.edu.cn,winsty@gmail.com,zhang-lq@cs.sjtu.edu.cn}

%\author{Jianfu Zhang\\
%Shanghai Jiao Tong University\thanks{Representing Key Laboratory of Shanghai Education Commission for Intelligent Interaction and Cognitive Engineering, Department of Computer Science and Engineering, Shanghai Jiao Tong University.}\\
%{\tt\small c.sis@sjtu.edu.cn}
%\and
%Naiyan Wang\\
%Tusimple\\
%{\tt\small winsty@gmail.com}
%\and
%Liqing Zhang\\
%Shanghai Jiao Tong University\\
%{\tt\small zhang-lq@cs.sjtu.edu.cn}
%}
\maketitle
%\thispagestyle{empty}

%%%%%%%%% ABSTRACT

\begin{abstract}
%Variations like blurs, illumination conditions, persons' orientations or poses make aggregating information from surveillance camera image sequences a very important issue in video based re-identification problems. Existing methods mostly use some trivial solutions like using the mean of all the features from different images of a person which suffer deeply from the large variance among the images. Recently, models based on recurrent neural networks and attention selection attempt to pick out the discriminative frames. However it costs to much and also too difficult to learn and integrate the saliency information from few image frames. In this paper, we proposed a model that treats video based re-identification as a set to set identification problem. We used reinforcement learning(RL), which is proved as a powerful tool to learn policies based on trial and error, to learn and select which images are in good quality to be aggregated.
Multi-shot pedestrian re-identification problem is at the core of surveillance video analysis. It matches two tracks of pedestrians from different cameras. In contrary to existing works that aggregate single frames features by time series model such as recurrent neural network, in this paper, we propose an interpretable reinforcement learning based approach to this problem. Particularly, we train an agent to verify a pair of images at each time. The agent could choose to output the result (same or different) or request another pair of images to verify (unsure). By this way, our model implicitly learns the difficulty of image pairs, and postpone the decision when the model does not accumulate enough evidence. Moreover, by adjusting the reward for unsure action, we can easily trade off between speed and accuracy. In three open benchmarks, our method are competitive with the state-of-the-art methods while only using 3\% to 6\% images. These promising results demonstrate that our method is favorable in both efficiency and performance.
\end{abstract}

%%%%%%%%% BODY TEXT
%-------------------------------------------------------------------------
\section{Introduction}
Pedestrian Re-identification (re-id) aims at matching pedestrians in different tracks from multiple cameras. It helps to recover the trajectory of a certain person in a broad area across different non-overlapping cameras. Thus, it is a fundamental task in a wide range of applications such as video surveillance for security and sports video analysis. The most popular setting for this task is single shot re-id, which judges whether two persons at different video frames are the same one. This setting has been extensively studied in recent years\cite{hongyang, AhmedJM15, LiZXW14, Wang_2016_CVPR, ijcai2017-305}. On the other hand, multi-shot re-id (or a more strict setting, video based re-id) is a more realistic setting in practice, however it is still at its early age compared with single shot re-id task.

Currently, the main stream of solving multi-shot re-id task is first to extract features from single frames, and then aggregate these image level features. Consequently, the key lies in how to leverage the rich yet possibly redundant and noisy information resided in multiple frames to build track level features from image level features. A common choice is pooling\cite{mars} or bag of words\cite{ZhengSTWWT15}. Furthermore, if the input tracks are videos (namely, the temporal order of frames is preserved), optical flow\cite{Chung_2017_ICCV} or recurrent neural network (RNN)\cite{McLaughlinRM16,Zhou_2017_CVPR} are commonly adopted to utilize the motion cues.
%Since videos may contain richer but also more redundant and unfocused information than images, arbitrary length, different frame-rates, occlusions and noises make how to aggregate information from each video the most important problem for video-based re-id task.
However, most of these methods have two main problems: the first one is that it is computationally inefficient to use all the frames in each track due to the redundancy. The second one is there could be noisy frames caused by occlusion, blur or incorrect detections. These noisy frames may significantly deteriorate the performance.

\begin{figure}[t]
\begin{center}
   \includegraphics[width=1.0\linewidth]{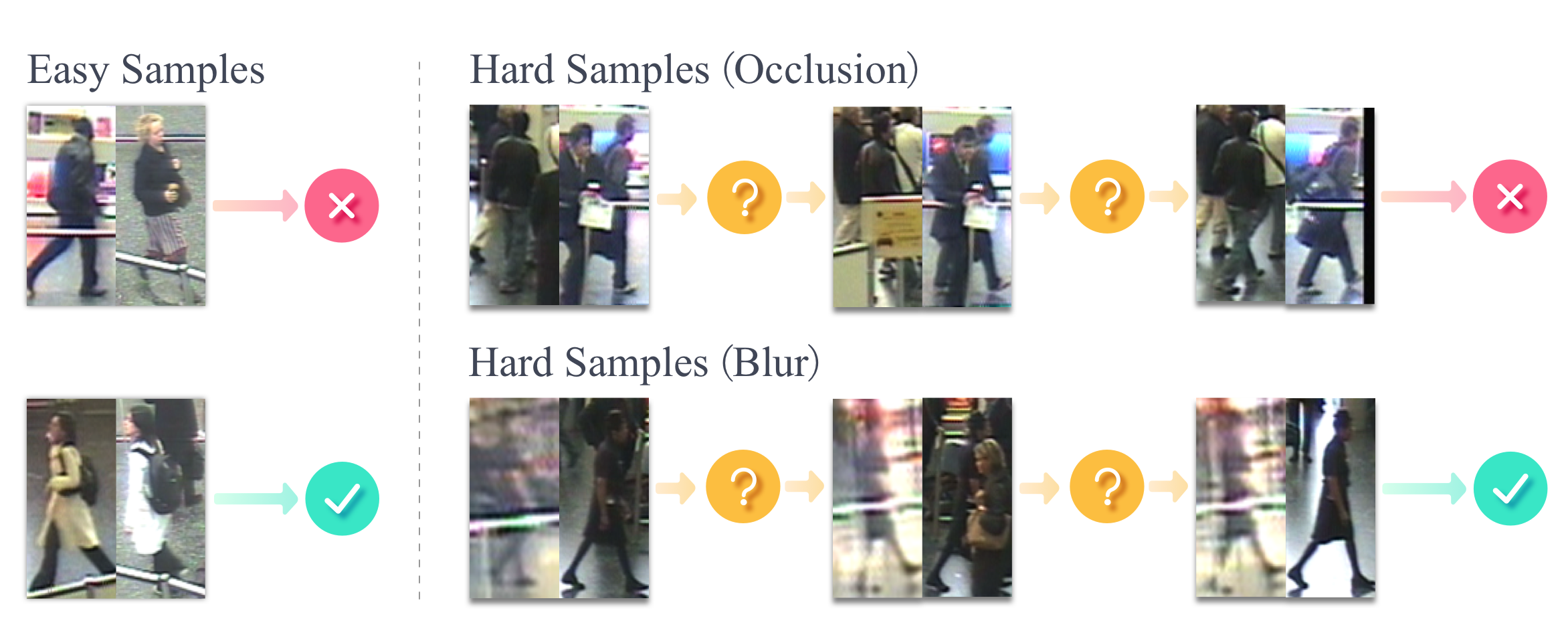}
\end{center}
   \caption{Examples to demonstrate the motivation of our work. For most tracks, several even only one pair of images are enough to make confident prediction. However, in other hard cases, it is necessary to use more pairs to alleviate the influence of these samples of bad quality.}
\label{fig:examples}
\end{figure}

To solve the aforementioned problems, we formulate multi-shot re-id problem as a sequential decision making task. Intuitively, if the agent is confident enough about existing evidences, it could output the result immediately. Otherwise, it needs to ask for another pair to verify. To model such human like decision process, we feed a pair of images from the two tracks to a verification agent at each time step. Then, the agent could output one of three actions: {\it same}, {\it different} or {\it unsure}. By adjusting the rewards of these three actions, we could trade off between the number of images used and final accuracy. We depict several examples in Fig.~\ref{fig:examples}. In case of easy examples, the agent could decide using only one pair of images, while when the cases are hard, the agent chooses to see more pairs to accumulate evidences. In contrast to previous works that explicitly deduplicate redundant frames\cite{DasPR17} or distinguish high quality from low quality frames\cite{LiuYO17}, our method could implicitly consider these factors in a data driven end-to-end manner. Moreover, our method is general enough to accommodate all single shot re-id methods as image level feature extractor even those non-deep learning based methods.%The whole process is more like a human style of doing verification. The agent verify the query identity and the gallery identity by comparing image pairs. When it is difficult to judge by the pairs already received for the agent, it will request for another pair.

The main contributions of our work are listed as following:
\begin{itemize}
	\item We are the first to introduce reinforcement learning into multi-shot re-id problem. We train an agent to either output results or request to see more samples. Thus, the agent could early stop or postpone the decision as needed. Thanks to this behavior, we could balance speed and accuracy by only adjusting the rewards.
	\item We verify the effectiveness and efficiency on three popular multi-shot re-id datasets. Along with the deliberately designed image feature extractor, our method could outperform the state-of-the-art methods while only using 3\% to 6\% images without resorting to other post-processing or additional metric learning methods.
	\item We empirically demonstrate that the Q function could implicitly indicate the difficulties of samples. This desirable property makes the results of our method more interpretable.
    %\item Our approach provides a faster decision process using less information by adding a negative reward when the agent seek for the next pair of images during the verification or it has already received enough information. We use only $?\%$ of the frames of the video to make the decision.
    %\item By postponing decisions, we minimize the effect of the frames which are in poor quality or complicate for the agent. To maximize the summed reward, the agent will eventually find the best aggregation of the images.
\end{itemize}

%-------------------------------------------------------------------------
\section{Related Work}
Pedestrian re-identification for single still images has been explored extensively in these years. These researches mainly focused on two aspects: the first one is to extract features that are both invariant and discriminative from different viewpoints to overcome difficulties such as illumination changes, occlusions, blurs, etc. Representative works before deep learning age include \cite{xwgd07,ikaa13, ranksvm}. However, these hand-crafted features are subverted by the rapidly developed Convolutional Neural Networks (CNN) in recent years. CNN has become \emph{de facto} standard for feature extraction.
The second aspect is metric learning. Metric learning embeds each sample into a latent space that preserves certain relationships of samples. Popular methods including Mahalanobis distance metric (RCA)\cite{rca}, Locally Adaptive Decision Function (LADF)\cite{ladf} and Large Margin Nearest Neighbor (LMNN)\cite{lmnn}.

These two streams have met in the deep learning age: Numerous work focus on learning discriminative features by the guide of metric learning based loss funcions.
The earliest work was proposed by Chopra \etal in \cite{ChopraHL05}. They presented a Siamese architecture to learn similarity for face verification task with CNN. Schroff \etal proposed triplet loss in FaceNet~\cite{SchroffKP15} to learn discriminative embeddings by maximizing the relative distance between matched pairs and mismatched pairs. Inspired by these methods for face verification, deep learning methods for image based re-identification have also shown great progress in recent years\cite{hongyang, LiZXW14, AhmedJM15}. Recently, \cite{Zhao_2017_CVPR, Zhao_2017_ICCV} utilized domain knowledge to improve performance: They incorporated pedestrian landmarks to handle body part misalignment problem. Concurrently, many deep learning based multi-task methods are proposed and reported promising performance. Wang \etal \cite{Wang_2016_CVPR} proposed a joint learning framework by combining patch matching and metric learning. Li \etal \cite{ijcai2017-305} proposed a multi-loss model combining metric learning and global classification to discover both local and global features.

%\footnote{Deep learning based single image re-id, this section needs to be largely expanded. Cite more works start from siamese net to triplet. They are all needed. Rewrite this paragraph.}Deep learning methods\cite{hongyang, rhsc06, cwcx17} has shown a great progress in image based re-identification recent years, mainly using neural network framework to do metric learning. Recently, many multi-task methods are proposed based on deep learning, \cite{Wang_2016_CVPR} combine the matching of single-image representation and the classification of cross-image representation using convolutional neural network(CNN).
\begin{figure*}[!htb]
\begin{center}
   \includegraphics[width=0.9\linewidth]{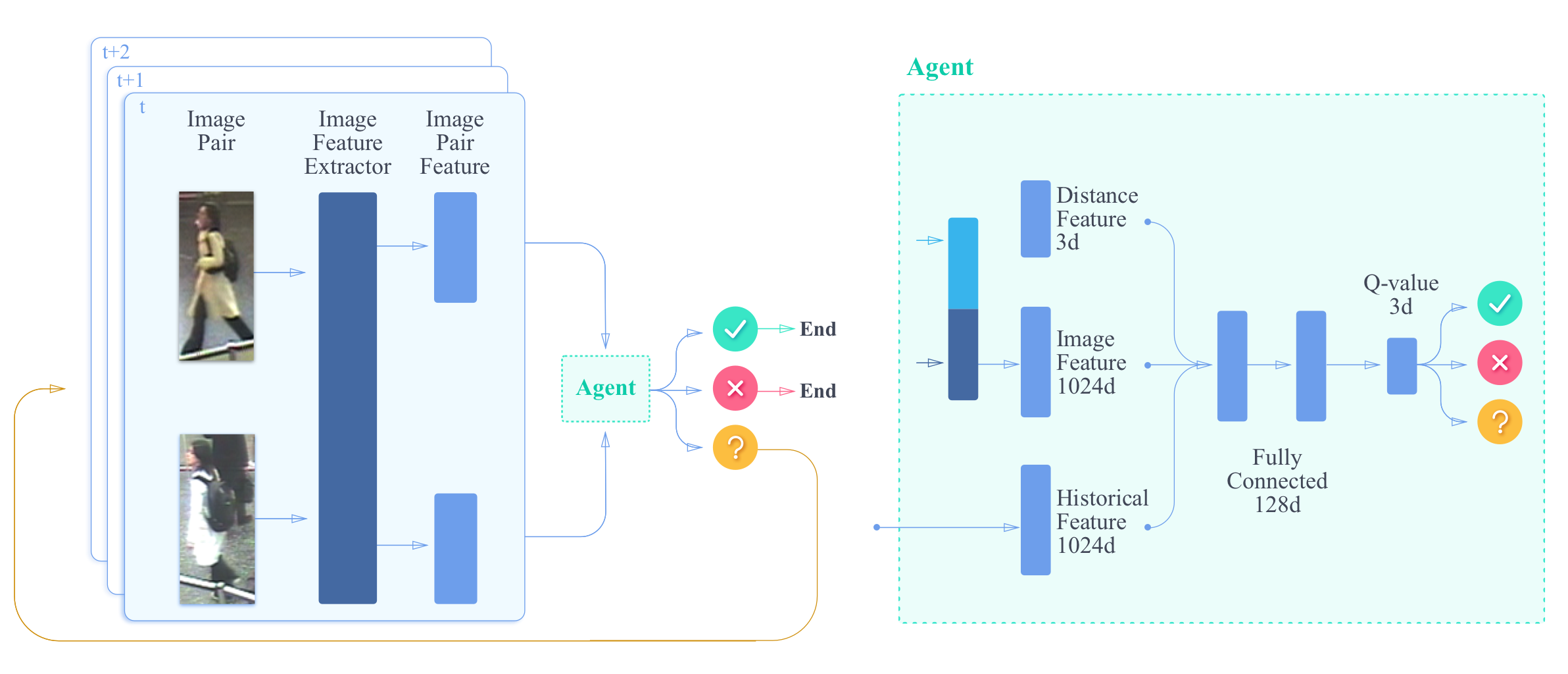}
\end{center}
   \caption{An illustration of our proposed method. Firstly we train an image level feature extractor (the left part) and then aggregate sequence level feature with an agent (the right part). The agent takes several kinds of features of one pair of images, and take one of three possible actions. If the taken action is ``unsure", the above process is repeated again.}
\label{fig:structure}
\end{figure*}

Compared with image based re-id task, multi-shot re-id problem is a more realistic setting, since the most popular application of re-id problem is surveillance video. It at least provides several representative frames after condensation, or even the entire videos are stored. Consequently, how to utilize such multi-frame information is at the core of multi-shot re-id task. Flow Energy Profile\cite{LiuMZH15} is proposed to detect walking cycles with flow energy profile to extract spatial and temporal invariant features. In \cite{ZhengSTWWT15}, Bag-of-words are adopted with learned frame-wised features to generate a global feature. Not surprisingly, deep learning also expressed its power in multi-shot re-id problem. A natural choice for temporal model in deep learning is Recurrent Neural Network (RNN). In the pioneering work \cite{McLaughlinRM16}, McLaughlin \etal first extracted features with CNN from images and then use RNN and temporal pooling to aggregate those features. Similarly, Chung \etal \cite{Chung_2017_ICCV} presented a two stream Siamese network with RNN and temporal pooling for each stream. Recently, this idea was extended with spatial and temporal attention in \cite{Zhou_2017_CVPR, abs-1708-02286} to automatically pick out discriminative frames and integrate context information. Another interesting work is \cite{LiuYO17}. In \cite{LiuYO17}, a CNN model learns the quality for each image, and then the video is aggregated with the image features weighted by the quality.

The goal of Reinforcement Learning (RL) is to learn policies based on trial and error in a dynamic environment. In contrast to traditional supervised learning, reinforcement learning trains an agent by maximizing the accumulated reward from environment. Additional to its traditional applications in control and robotics, recently RL has been successfully applied to a few computer vision tasks by treating them as a decision making process\cite{CaicedoL15,MalmirSFFMC17,Huang_2017_ICCV,LanWGZ17,KarayevFD14,MathePS16}. Some closely related works include: In \cite{Huang_2017_ICCV}, the features for visual tracking problem are sorted by their costs, and then an agent is trained to decide whether current features are good enough to make accurate prediction. If not, it proceeds to the next feature. By this way, the agent saves unnecessary computation of expensive features. %In \cite{LanWGZ17}, RL is applied to learn attention for detected bounding boxes by iteratively detecting and removing irrelevant pixels in each image.
\cite{KarayevFD14,MathePS16} are two works which applied RL techniques to object detection task.
In \cite{KarayevFD14}, the authors aimed to solve this task by limited budget which can be wall time, computing resources or etc. An agent is trained to learn a sequential policy for feature selection and stop before the cost budget is exhausted. While in \cite{MathePS16} an agent is trained to learn whether to sample more image regions for better accuracy or stop the search. Our method shares the same spirit with these works, but tailored for multi-shot re-id problem.

%-------------------------------------------------------------------------
\section{Method}
In this section, we will introduce our approach to multi-shot re-id problem. First, we will start with a formal formulation of this problem, and then present each component of our method. The overview of our method is depicted in Figure ~\ref{fig:structure}.

\subsection{Formulation}
%In multi-shot re-id task, we are given two sequences $(\MX, \MY)=(\{x_1, \dots,  x_n\}, \{y_1, \dots, y_n\})$, where $x_1$ represents the first image in $\MX$. We compute the distance (or similarity) of these two sequences by:
%\begin{equation}
%  D(\MX, \MY) = \|g(f(x_1), \dots, f(x_n))-g(f(y_1), \dots, f(y_n)) \|_2 ^2.
%\end{equation}
%Here $f(\cdot)$ is a feature extractor that extracts discriminative feature for each frame, and $g(\cdot)$ is an aggregation function that aggregates image level features to sequence level feature.
%We then use it to rank all the queries.

In multi-shot re-id task, for each sequence in query identities, the goal is to rank all the gallery identities according to their similarities with the query identity. Given two sequences $(\MX, \MY)=(\{x_1, \dots,  x_m\}, \{y_1, \dots, y_n\})$, where $x$ and $y$ represent the images in $\MX$ and $\MY$, respectively. Let $f(x)$ be a feature extractor that extracts discriminative features for each image $x$, and $g(\MX)$ be an aggregation function that aggregates image level features of $\MX$ to sequence level feature. A similarity function $l(\cdot, \cdot)$ is designed to calculate the similarity between the query identity and gallery identity. According to the similarity computed by $l(\cdot, \cdot)$, we sort all the gallery identities for each query identity.

In the sequel, we will first present the details of our single image feature extractor $f(\cdot)$ in Sec.~\ref{sec:singlereid}. It is built with a CNN trained with three different loss functions. Next, we elaborate our reinforcement learning based aggregation method $g(\cdot)$ and $l(\cdot, \cdot)$ in Sec.~\ref{sec:rlreid}.

%Our model contains two parts: the first part is a image based Re-ID discrimination network by training a convolutional neural network(CNN) to learn a metric embedding in a multi-task learning framework. For the second part, two sets of images are provided and each of them are from the same identity. Our goal is to learn a policy by reinforcement learning to judge whether the given two sets of images are from the same identity or not, using the embedding generated by the first part.

\subsection{Image Level Feature Extraction}
\label{sec:singlereid}
For single image feature extractor, a CNN is trained to embed an image into a latent space that preserves certain relationships of samples. To achieve this goal, we train a CNN with combination of three different kinds of loss functions: classification loss, pairwise verification loss~\cite{ChopraHL05} and triplet verification loss~\cite{SchroffKP15}. According to a recent work~\cite{mcvpr}, multiple loss functions could better ensure the structure of the latent space and margins between samples. Particularly, we optimize large margin softmax loss\cite{lsoftmax} instead of softmax loss, since it demonstrates extraordinary performance in various classification and verification tasks.
\paragraph
{\bf Implementation details:} We use two well-known network structures Inception-BN\cite{incptbn} and AlexNet~\cite{imagenet} pre-trained on ILSVRC classification dataset\cite{imagenet} as base networks. We choose these two networks with different capacity and expression power to demonstrate the universality of our proposed aggregation method. In specific, we use the features from the last pooling layer as image level features. In training, we set the margin in triplet loss to $0.9$. For large margin softmax, we set $\beta=1000$, $\beta_{min}=3$, and the margin as $3$. For more details of these parameters, please refer to~\cite{lsoftmax}.
We optimize the network by momentum SGD optimizer with $320000$ iterations. The learning rate is $0.01$ and multiplied by $0.1$ after $50000$ and $75000$ iterations, respectively.

As an important baseline, we simply use the average of $l_2$-normalized features from all the images as the feature for a sequence. Namely, the aggregation and similarity function is defined as:
\begin{equation} \label{eq:1}
  g(\MX)=\sum_i^m \frac{f(x_i)}{m}, \quad l(g(\MX), g(\MY)) = g(\MX)\cdot g(\MY)
\end{equation}
$\cdot$ representing inner product for two vectors. We rank all the gallery identities according to the value generated by $l(\cdot, \cdot)$.

\subsection{Sequence Level Feature Aggregation}
\label{sec:rlreid}
We formulate this problem as a Markov Decision Processes (MDP), described by $(\mathcal{S,A,T,R})$ as the states, actions, transitions and rewards. Each time step $t$, the agent will get a selected image pair from the two input sequences to observe a state $s_t\in \mathcal{S}$ and then choose an action $a_t \in \mathcal{A}$ from the experience it has learned. Next the agent will earn a reward $r_t \in \mathcal{R}$ from the environment in training. After that if the episode is not terminated, the agent will receive another image pair determined by state transition distribution $\mathcal{T}(s_{t+1}|s_t, a_t)$ and turn to the next state $s_{t+1}$. We will elaborate the details of them in the sequel.
\paragraph
{\bf Actions and Transitions:} Initially, the agent is fed with an image pair selected from two selected sequences $\MX$ and $\MY$. Note that we don't assume the order of the input and \textbf{randomly} form the pair from two sequences. We have three actions for the agent: {\it same}, {\it different} and {\it unsure}. The first two actions will terminate the current episode, and output the result immediately. We anticipate when the agent has collected enough information and is confident to make the decision, it stops early to avoid unnecessary computation. If the agent chooses to take action {\it unsure}, we will feed the agent another image pair.\\
{\bf Rewards:} We define the rewards as follows:
\begin{enumerate}
	\item $+1$, if $a_t$ matches $gt$.
	\item $-1$, if $a_t$ differs from $gt$, or when $t=t_{max}$, $a_t$ is still {\it unsure}.
	\item $r_p$, if $t < t_{max}$, $a_t$ is {\it unsure}.
\end{enumerate}
%\begin{equation}
%    R(s_t)=
%    \begin{cases}
%        +1 &\mbox{if $s_t$ is terminal and $a_t=gt$}\\
%        -1 &\mbox{if $s_t$ is terminal and $a_t \neq gt$ or $t=t_{max}$}\\
%        r_p &\mbox{if $s_t$ is not terminal and $t < t_{max} $}
%    \end{cases}
%\end{equation}
Here $t_{max}$ is defined as the maximum time step for each episode. $gt$ is the ground truth. $r_p$ is defined as a penalty (negative reward) or reward for the agent seeking for another image pair. If $r_p$ is negative, it will be penalized for requesting more pairs; on the other hand, if $r_p$ is positive, we encourage the agent to gather more pairs, and stop gathering when it has collected $t_{max}$ pairs to avoid a penalty of $-1$. The value of $r_p$ may strongly affect the agent's behavior. We will discuss its impact in Sec.~\ref{ref:ablation}.
\paragraph
{\bf States and Deep Q-learning:}
We use Deep Q-Learning~\cite{DQN} to find the optimal policy. For each state and action $(s_t, a_t)$, $Q(s_t, a_t)$ represents the  discounted accumulated rewards for the state and action. In training, we could iteratively update the Q function by:
\begin{equation}
  Q(s_t,a_t)=r_t+\gamma \mathop{\max}_{a_{t+1}} Q(s_{t+1}, a_{t+1}).
\end{equation}
The state $s_t$ for time step $t$ in the episode consists of three parts. The first part is the observation $o_t$ which is composed of the image features of current pair $(f(x), f(y))$ generated by the image feature extractor mentioned in Section~\ref{sec:singlereid}, which is defined as $o_t=|f(x_t)-f(y_t)|$.%\footnote{Why we need absolute value of this?? It should be a vector right? Also pay attention to the subscript. - Because of symmetry: we don't want to get different results for input (x, y) and (y, x). So here most commonly used methods are absolute difference and product of each element of the vector. I tried product and it's 1 to 2 points worse compared with abs diff. I also did an experiment if I remove abs here and the network failed to converge.} 
The second part is a weighted average of the difference between historical image features of two sequences.
This part makes the agent be aware of the previous image pairs it has already seen before. In specific, for each observation $o_t$ the weight $w_t$ is defined as:\\
\begin{equation}
    w_t = 1.0 - \frac{e^{Q_u}}{e^{Q_s} + e^{Q_d} + e^{Q_u}}
\end{equation}
where $Q_u$ is short for $Q(s_t, a_t=\mathit{unsure})$, and vice versa.
The weight decreases as $Q_u$ increases, as higher $Q_u$ may indicate that current pair of images are hard to distinguish. The aggregated features should be affected as small as possible. As a result, $h_t$ is the weighted average of the historical features for $t>1$:
\begin{equation}
    h_t = \frac{\sum_{i=1}^{t-1} w_i\times o_i}{\sum_{i=1}^{t-1} w_i}.
\end{equation}
$h_t=o_t$ when $t=1$.
\footnote{Note that since $h_t=0$ implies $f(x)=f(y)$, it will introduce a strong bias to make the agent to choose ``same'' leading a poor performance if we set $h_t=0$ when $t=1$.}
Note that though the Q function is not specifically trained for sample weighting, it still reflects the importance of each frame. We leave end-to-end learning of the weights as our future work.

We also augment the image features with hand-crafted features for better discrimination. For each time step $t$, we calculate the distance $\| f(x_{i}) - f(y_j) \|_2 ^2$ for all $1\leq i,j < t$, and then add the maximum, minimum and mean of them to the input, which results in $3$ dimension extra features.
\footnote{Here we don't make time step $t$ as a part of the state-space. Since the feature extractor fits better in the training set, the agent uses less time steps to verify samples in training set compared with that in testing set, adding $t$ to the state-space will cause overfitting issues.}

The structure of the Q-network is shown in Fig.\ref{fig:structure}. We simply use a two layer fully connected network as the Q function. Each fully connected layer has $128$ outputs and is followed by an ReLU activation function.\\
{\bf Testing:} For each query video sequences we play one episode and take the difference of the Q-value of action {\it same} and {\it different} at the terminal step as the final ranking score. Note that the Q-net essentially combines aggregation function $g(\cdot)$ and similarity function $l(\cdot, \cdot)$.\\
% \begin{equation} \label{eq:2}
%   g(\MX, \MY)=Q_{s}-Q_{d}
% \end{equation}
% Then we rank the gallery identities by this value.\\
\noindent
{\bf Implementation details:} In training phase, for each episode we randomly choose positive or negative sequence pairs with ratio $1:1$. We feed the weighted historical features, features of current step and hand-crafted distance features into the Q-Net. The whole net along with the single image feature extractor is trained end-to-end except for fixing the first two stages of the base networks.

We train the Q-Net for 20 epochs by momentum SGD optimizer, 100000 iterations for each epoch. We use $\epsilon$-greedy learning\cite{epsilon} as the exploration strategy and anneal $\epsilon$ linearly from $1$ to $0.1$ in the first $10$ epochs. Learning rate is set to $0.0001$, discount factor $\gamma=0.9$ and batch size is $16$. Experience replay is used and the memory buffer size is set to  $5000$. It takes $5.502$ and $2.613$ ms per episode for Inception-BN and Alexnet to verify a single pair of sequences on a Maxwell Titan X GPU. All these runtimes include the time of both image level feature extractor and Q-Net.

%-------------------------------------------------------------------------
\begin{figure*}[!ht]
\begin{center}
   \includegraphics[width=0.3\linewidth]{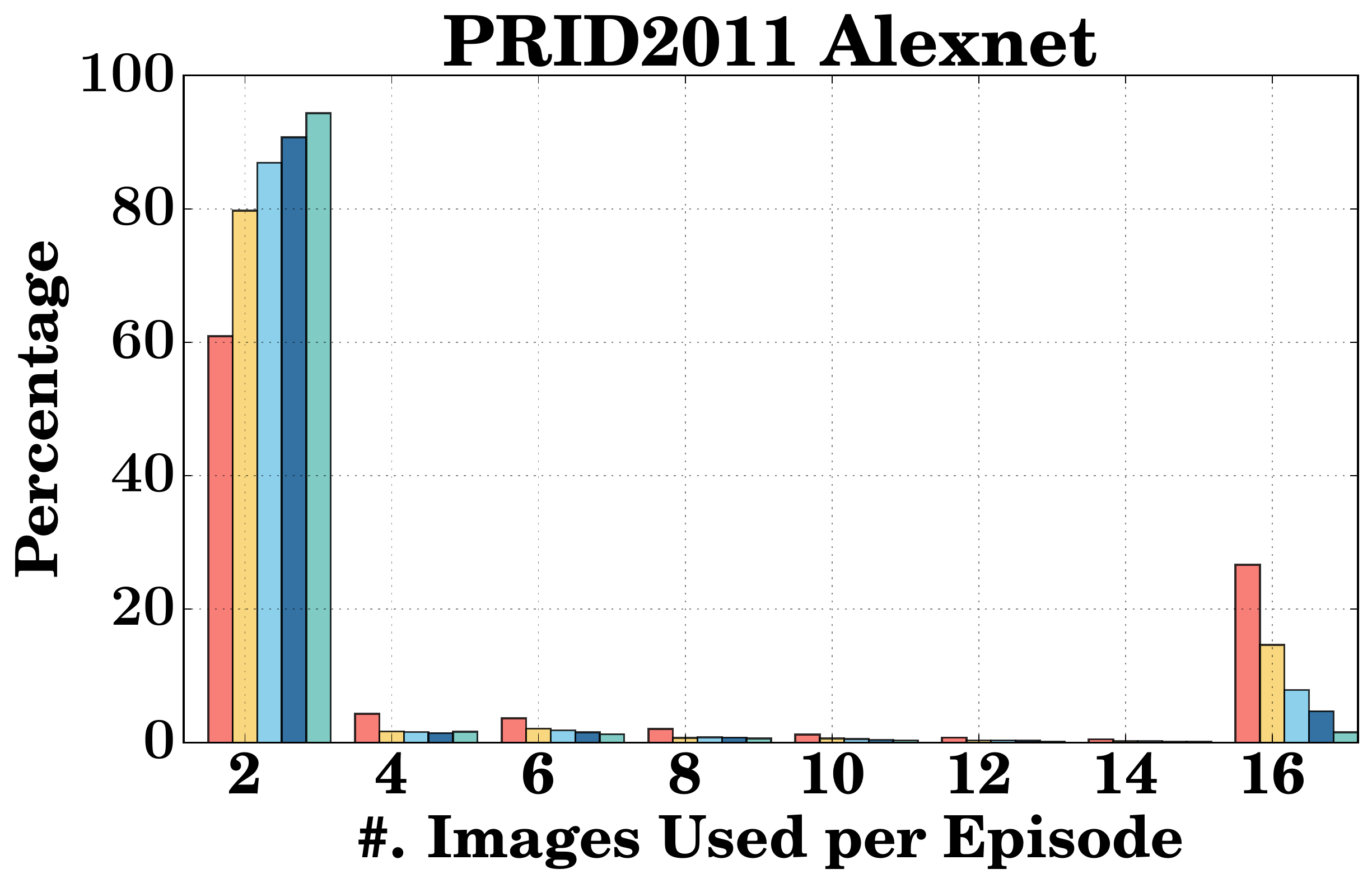}
   \includegraphics[width=0.3\linewidth]{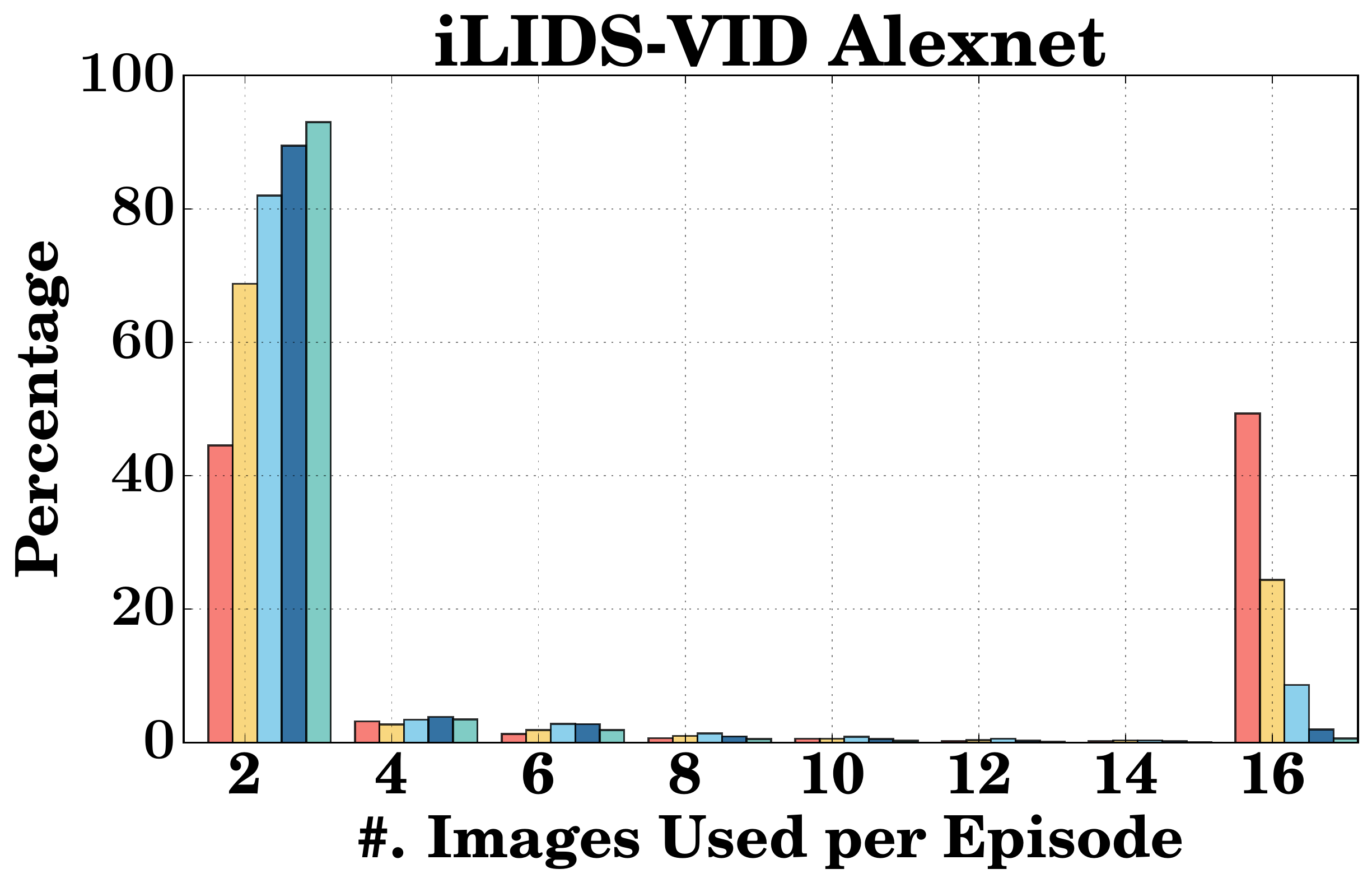}
   \includegraphics[width=0.3\linewidth]{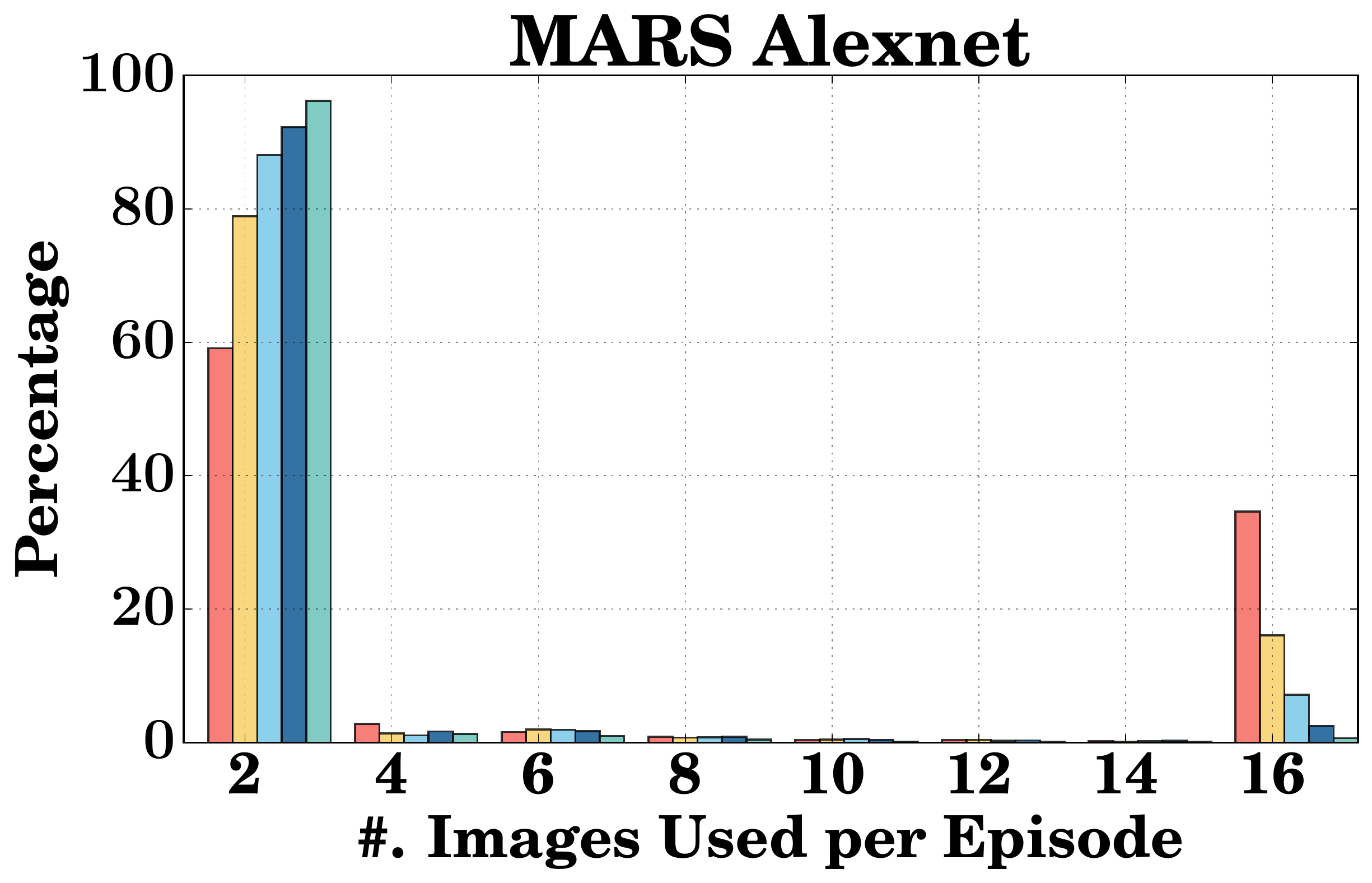}
   \includegraphics[width=0.3\linewidth]{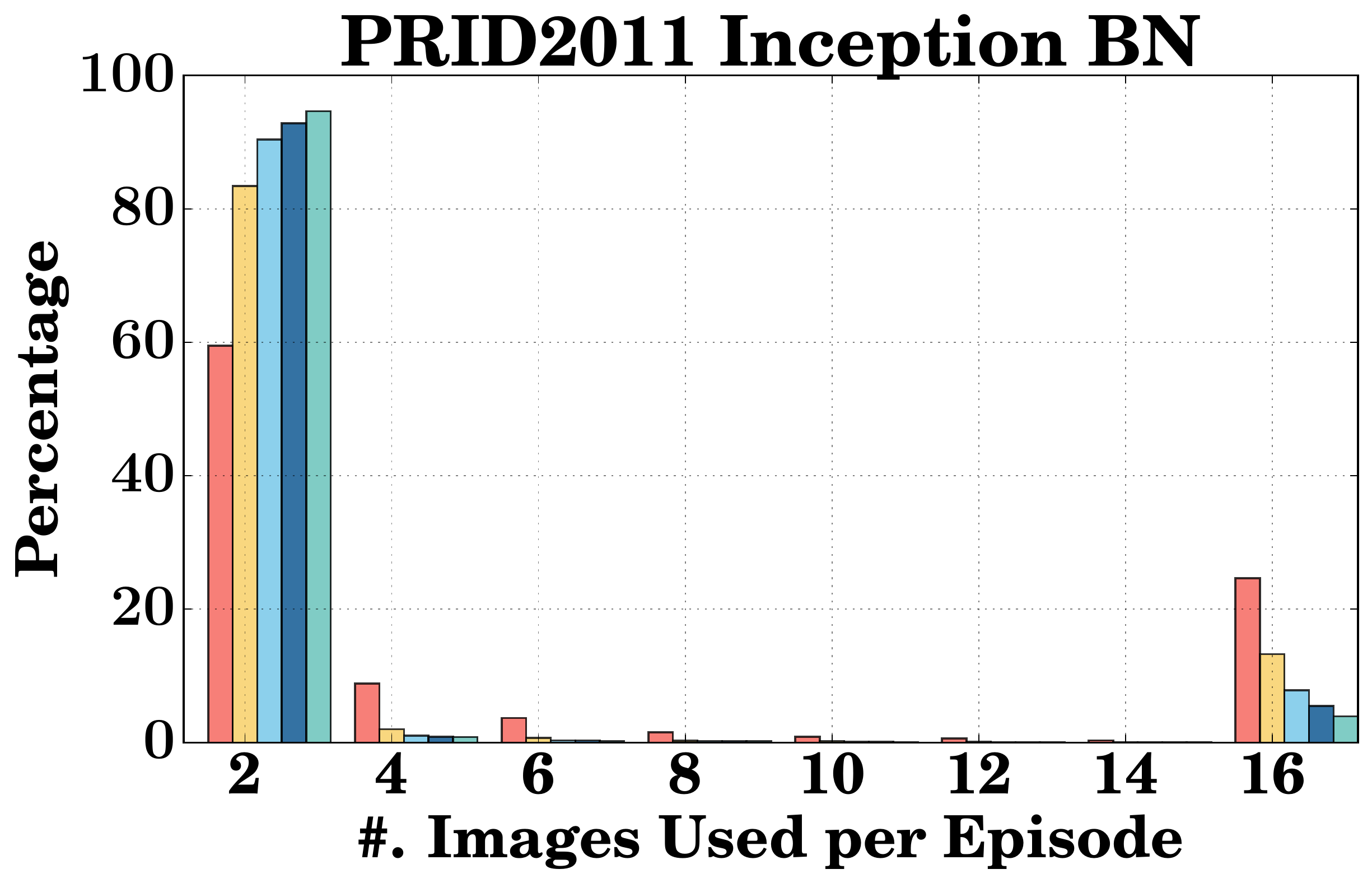}
   \includegraphics[width=0.3\linewidth]{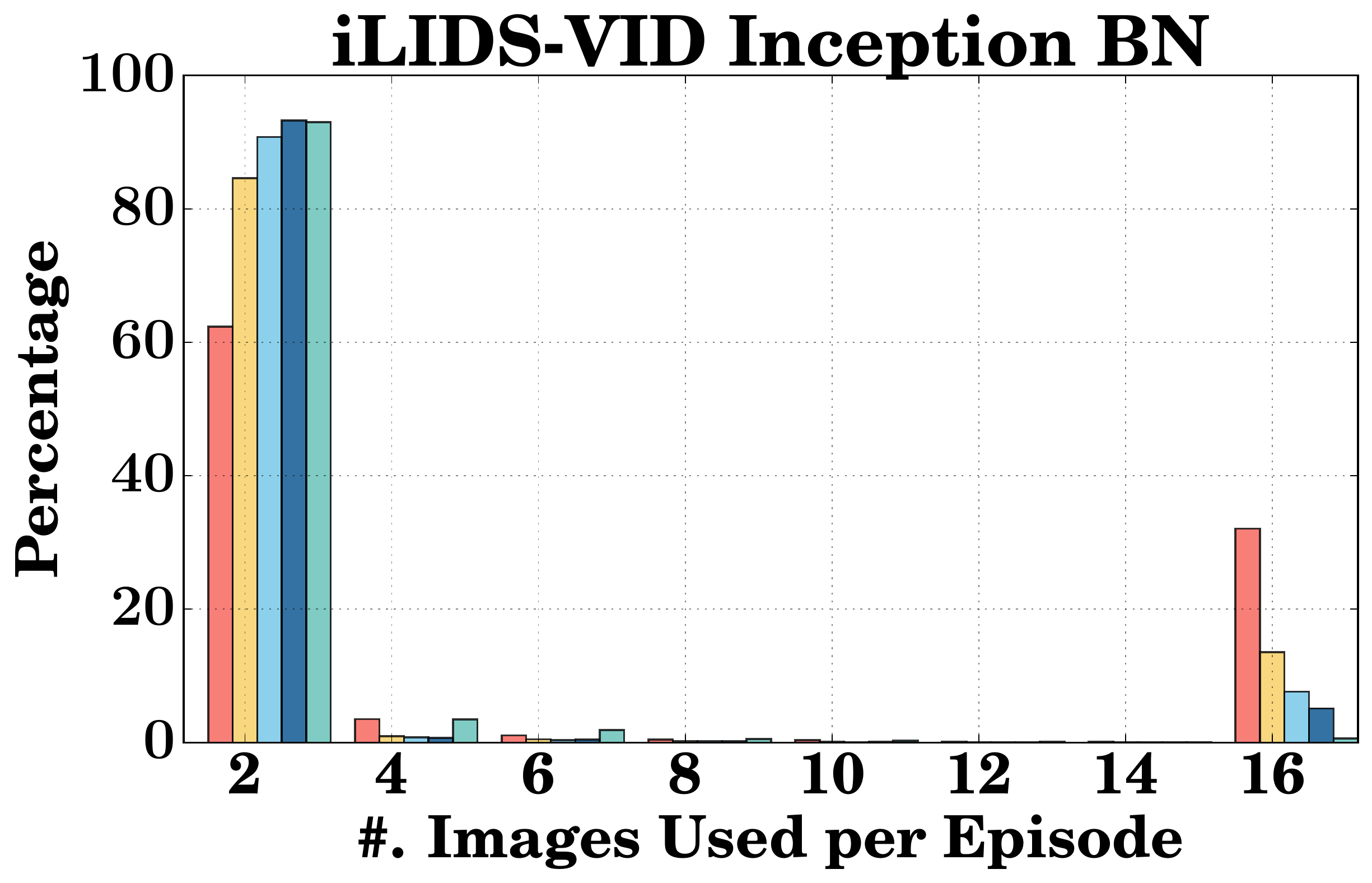}
   \includegraphics[width=0.3\linewidth]{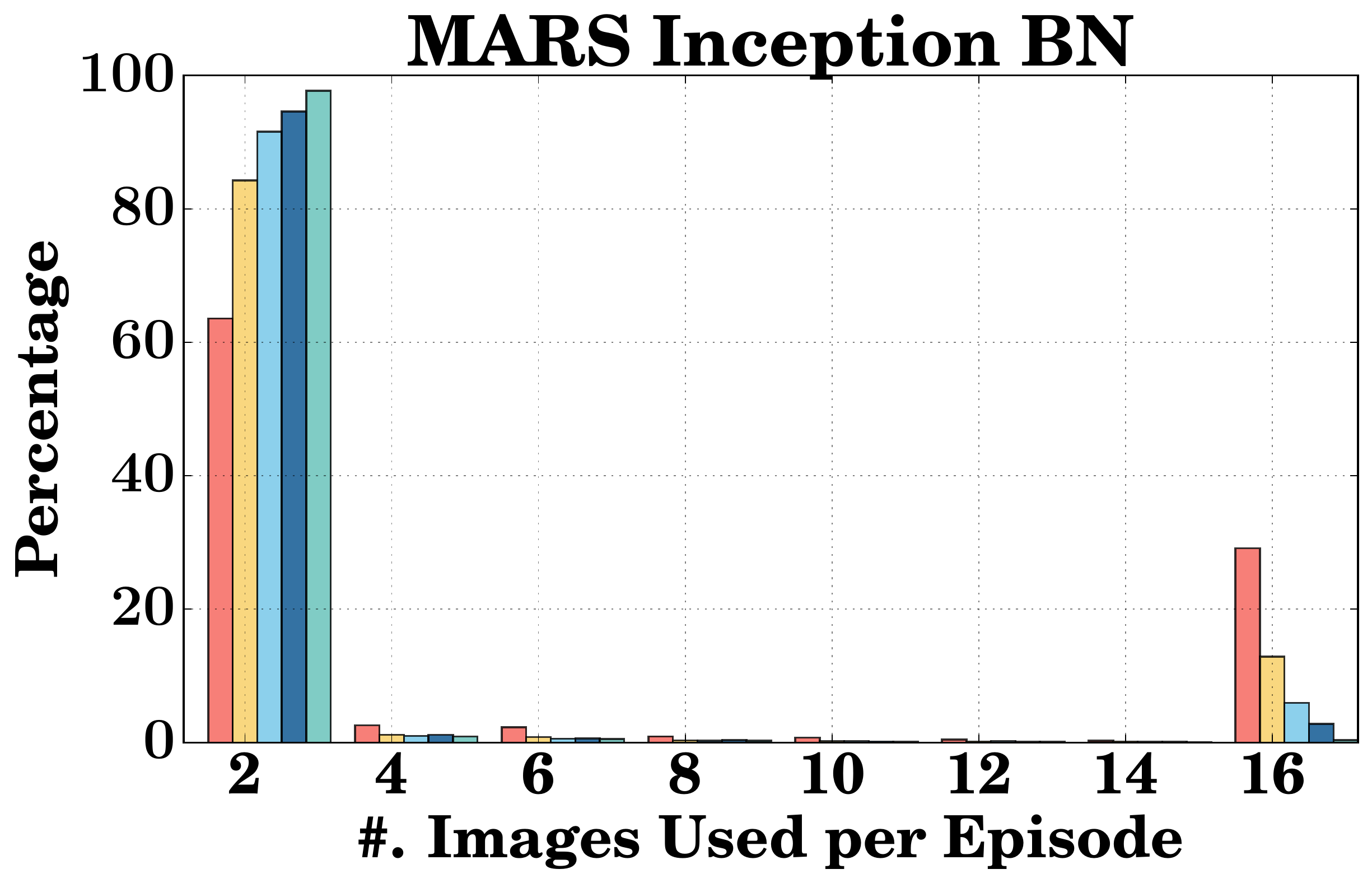}
   \includegraphics[width=0.64\linewidth]{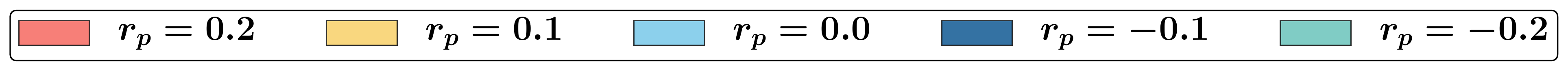}
\end{center}
   \caption{Statistics of the number of images used in each episode of our model with different reward for action {\it unsure}.}
\label{fig:episodelength}
\end{figure*}

\section{Experiments}
In this section, we will present the results of our method on three open benchmarks, and compare it with other state-of-the-art methods. We will first introduce the datasets and evaluation metric used, and then present the ablation analyses of our method. After comparisons with other methods, we will also present some qualitative results to interpret the mechanism of our methods.

%\subsection{Datasets} \footnote{If it is too long, we can remove this subsection, and merge it with the next subsection.}
%The iLIDS-VID dataset\cite{ilids} contains $300$ identities with $600$ image sequences from two cameras. The length for each image sequence ranges from $23$ to $192$ frames. The challenge of this dataset is mainly due to severe occlusion. The bounding boxes are human annotated.

%The PRID2011\cite{prid} dataset consists two cameras with $385$ identities in camera A and $749$ identities in camera B. $200$ identities appear in both camera. The length for each image sequence  varies from $5$ to $675$ frames. Same as iLIDS-VID, the bounding boxes are labeled by human.

%The Motion Analysis and Re-identification Set (MARS)\cite{mars} is a recently released large scale dataset containing $1261$ identities and $20715$ tracklets under $6$ different camera views. Bounding boxes are generated by GMMCP\cite{DehghanAS15} tracker and Deformable Part Model (DPM)\cite{FelzenszwalbGMR10} pedestrian detector, which makes it quite noisy yet close to real applications.

\begin{figure*}[!ht]
\begin{center}
   \includegraphics[width=0.3\linewidth]{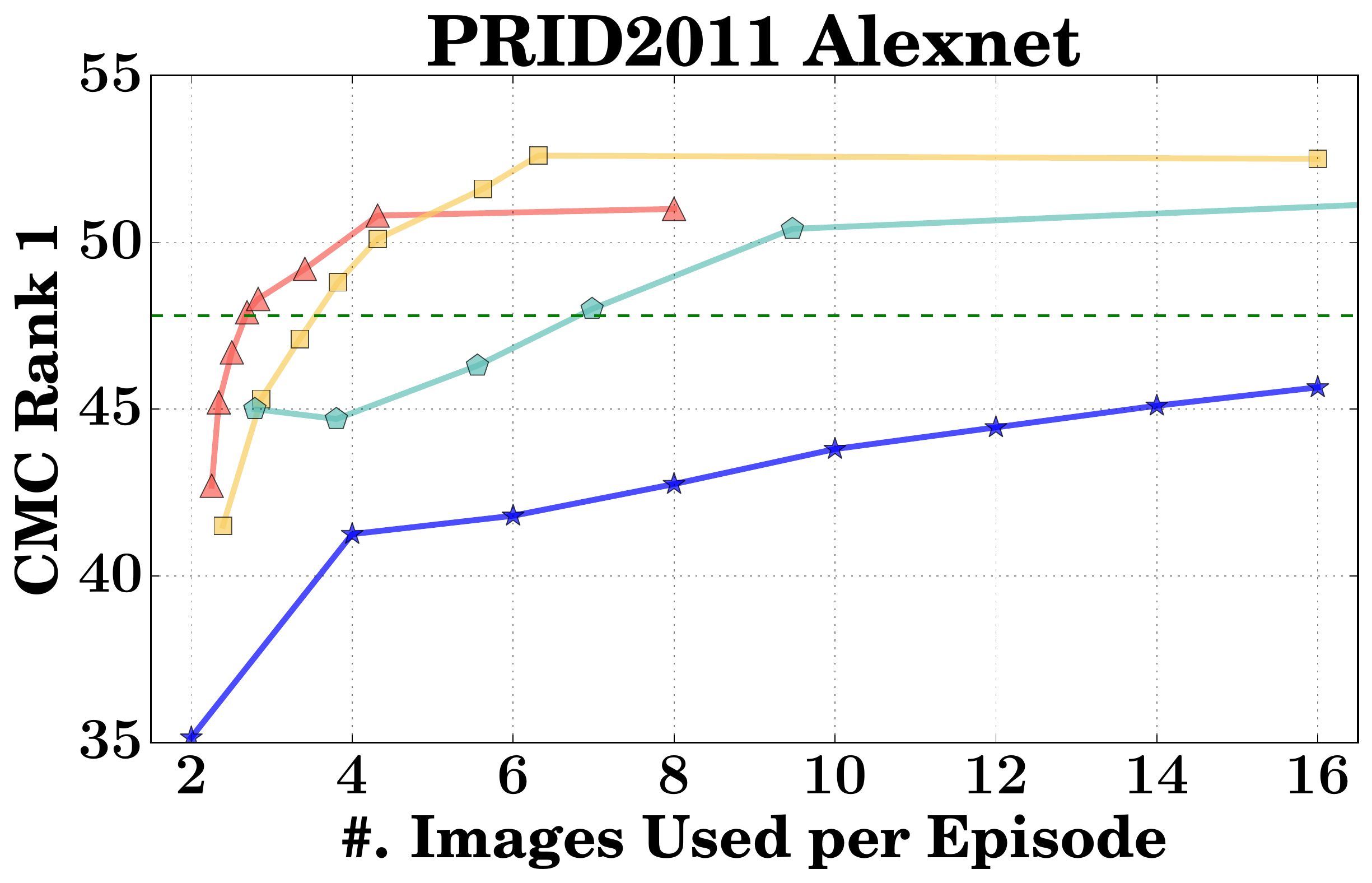}
   \includegraphics[width=0.3\linewidth]{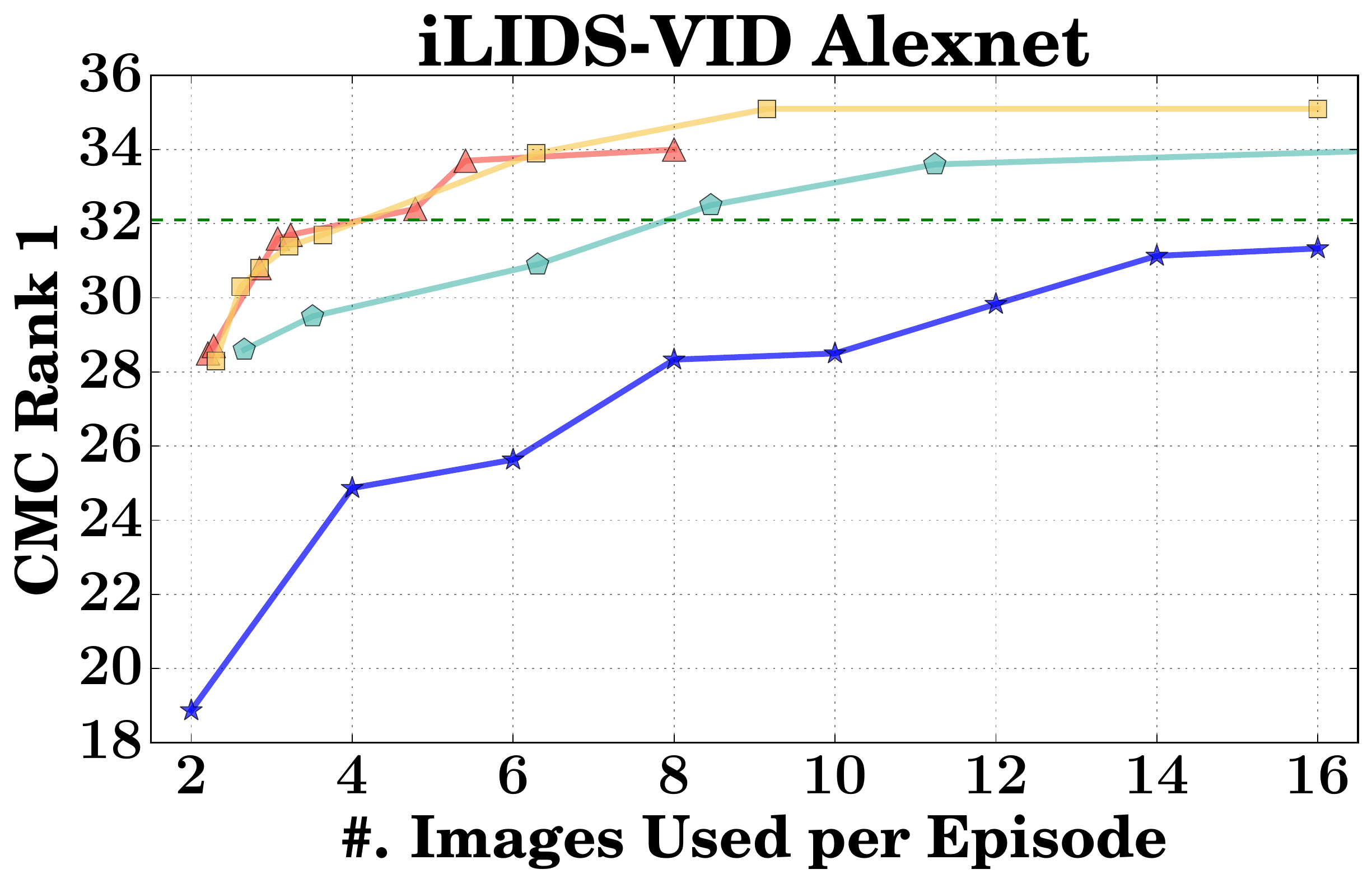}
   \includegraphics[width=0.3\linewidth]{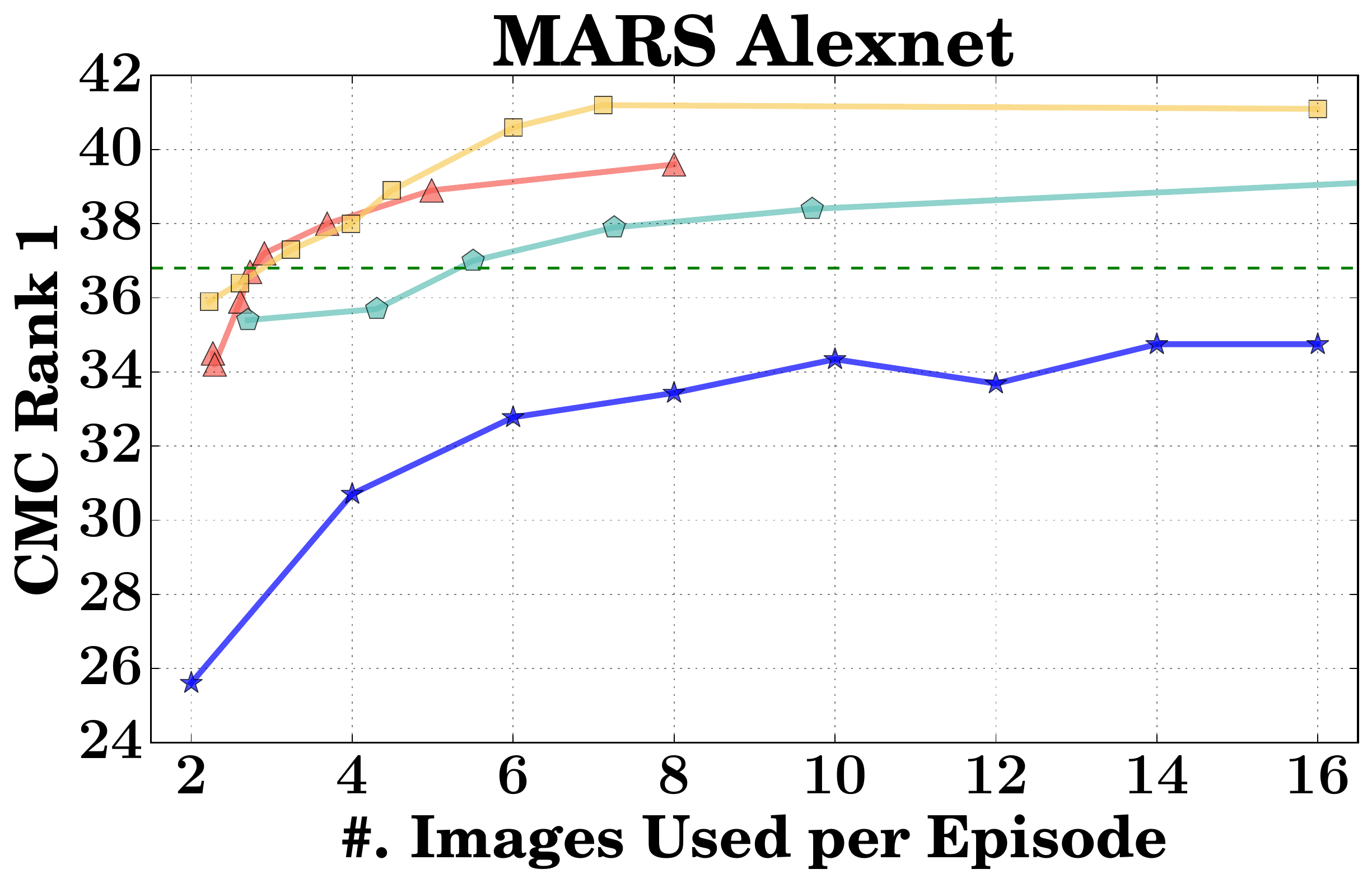}
   \includegraphics[width=0.3\linewidth]{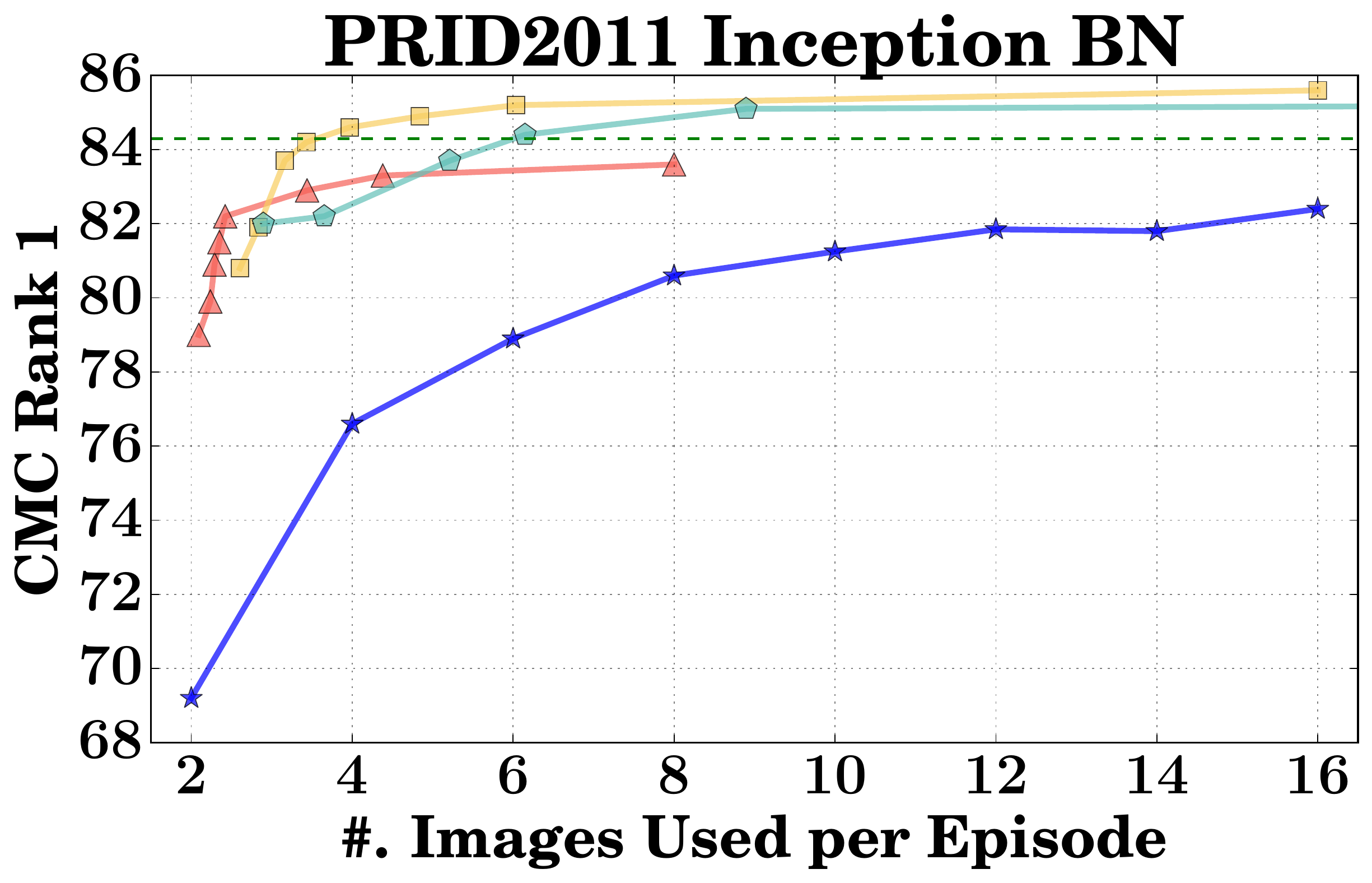}
   \includegraphics[width=0.3\linewidth]{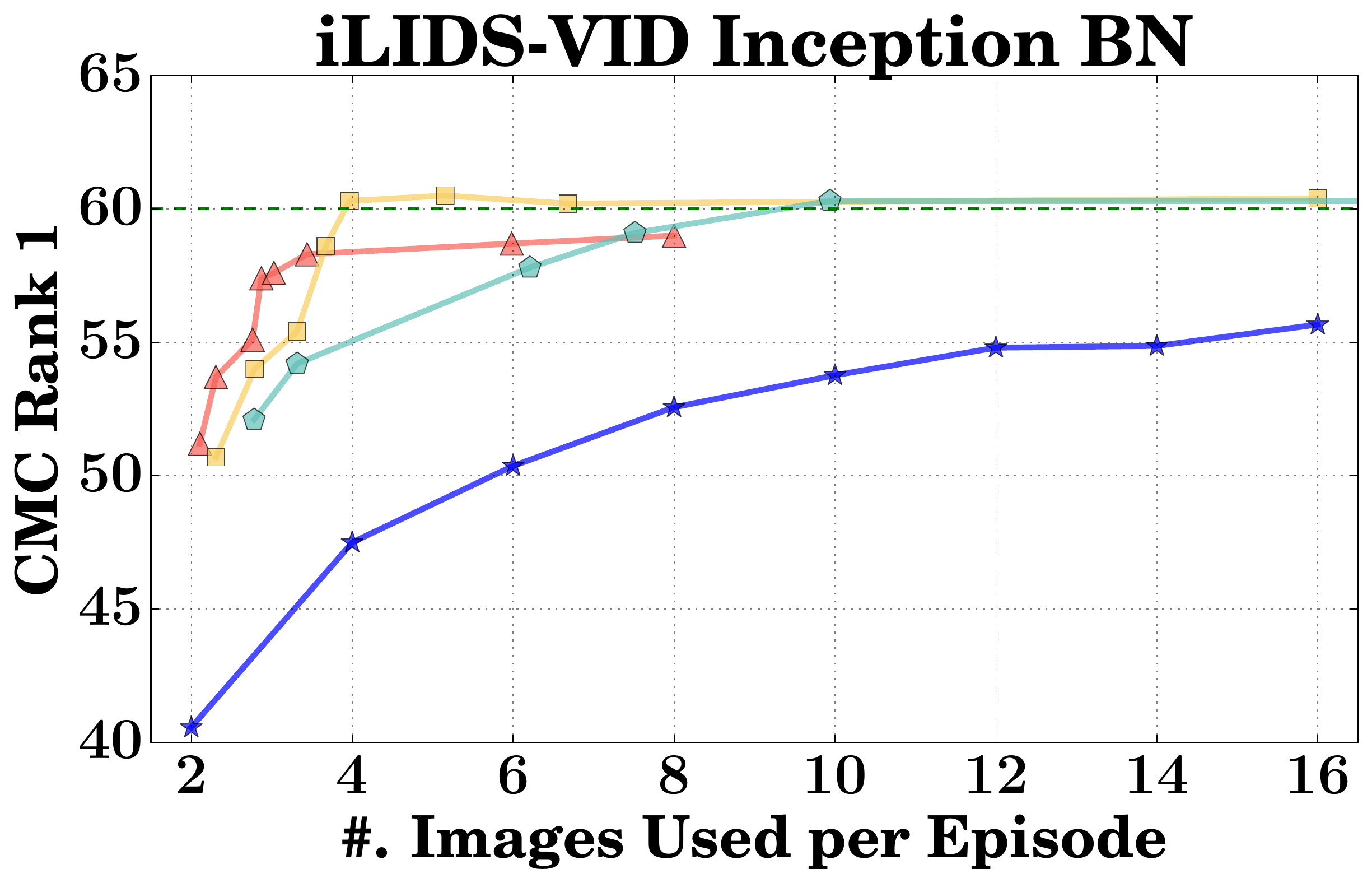}
   \includegraphics[width=0.3\linewidth]{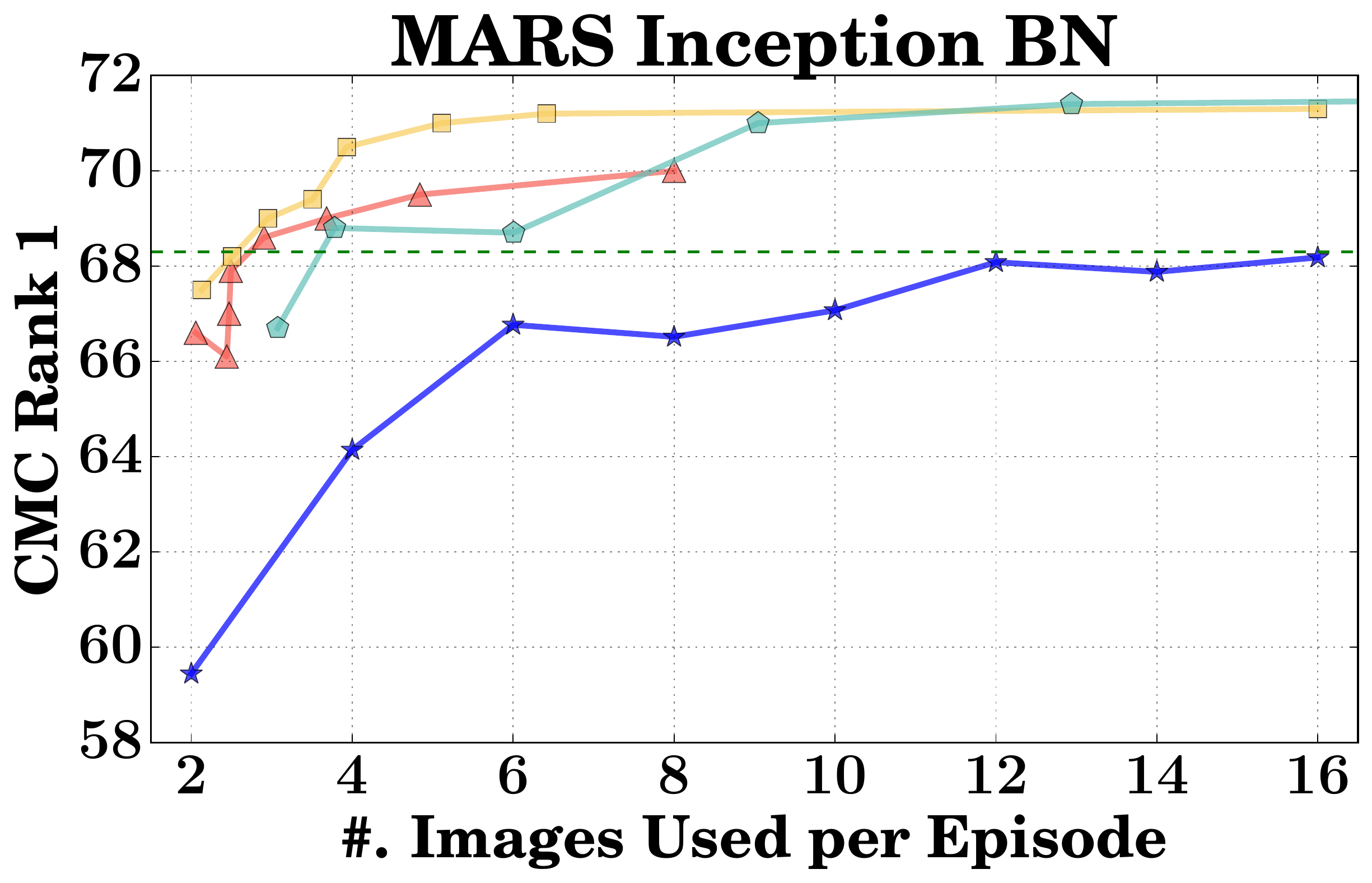}
   \includegraphics[width=0.64\linewidth]{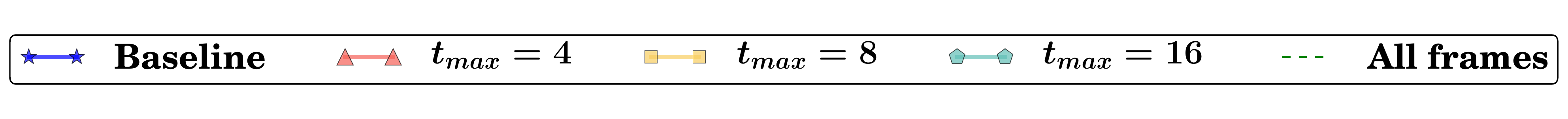}
\end{center}
	\vspace{-3mm}
   \caption{CMC Rank $1$ results for our model compared with baseline.}
\label{fig:c1curve}
\end{figure*}

\begin{table*}[!ht]
\begin{small}
\begin{center}
\begin{tabular}{l|cc|cc|cc}
\toprule
Dataset & \multicolumn{2}{|c|}{PRID2011} &  \multicolumn{2}{c|}{iLIDS-VID} & \multicolumn{2}{c}{MARS}\\
\hline
Settings & CMC1 & \#.of Images & CMC1 & \#.of Images & CMC1 & \#.of Images \\
\hline
All frames & 84.3 & 200.000 & 60.0 & 146.000 & 68.3 & 111.838 \\
$r_p=0.2$ & \textbf{85.2} & 6.035 & 60.2 & 6.681 & \textbf{71.2} & 6.417 \\
$r_p=0.1$ & 84.6 & 3.970 & \textbf{60.3} & 3.966 & 70.5 & 3.931 \\
$r_p=0$ & 83.7 & 3.162 & 55.4 & 3.134 & 69.0 & 2.952  \\
$r_p=-0.1$ & 81.9 & 2.835 & 54.0 & 2.789 & 68.2 & 2.507 \\
$r_p=-0.2$ & 80.8 & \textbf{2.605} & 50.7 & \textbf{2.307} & 67.5 & \textbf{2.130} \\
\hline
%$t_{max}=4$ & 82.2 & 2.461 & 57.6 & 3.027 & 68.6 & 2.904 \\
%$t_{max}=16$ & 85.3 & 32.000 & 60.0 & 32.000 & 71.7 & 32.000 \\
%$t_{max}=16, r_p=0.1$ & 85.1 & 8.894 & 60.3 & 9.937 & 71.0 & 9.045 \\
No handcrafted features & 83.5 & 5.679 & 57.8 & 5.934 & 69.2 & 6.103 \\
DRQN & 83.2 & 4.314 & 59.8 & 5.109 & 69.9 & 4.577 \\
Sequential & 84.1 & 7.549 & 59.7 & 7.021 & 70.5 & 6.591 \\
Video fine-tune & 84.7 & 16.000 & 60.2 & 16.000 & 70.7 & 16.000 \\
\bottomrule
\end{tabular}
\end{center}
\end{small}
\vspace{-2mm}
\caption{Test results for our model based on Inception BN image feature extractor.}
\label{tab:2}
\end{table*}

\begin{table*}[!ht]
\begin{small}
\begin{center}
\begin{tabular}{l|cc|cc|cc}
\toprule
Dataset & \multicolumn{2}{|c|}{PRID2011} &  \multicolumn{2}{c|}{iLIDS-VID} & \multicolumn{2}{c}{MARS}\\
\hline
Settings & CMC1 & \#.of Images & CMC1 & \#.of Images & CMC1 & \#.of Images \\
\hline
All frames & 47.8 & 200.000 & 32.1 & 146.000 & 36.8 & 111.838 \\
$r_p=0.2$ & \textbf{52.6} & 6.316 & \textbf{35.1} & 9.154 & \textbf{41.2} & 7.119 \\
$r_p=0.1$ & 50.1 & 4.317 & 33.3 & 5.722 & 38.9 & 4.491 \\
$r_p=0$ & 47.1 & 3.349 & 31.7 & 3.637 & 37.3 & 3.238  \\
$r_p=-0.1$ & 45.3 & 2.870 & 30.3 & 2.614 & 36.4 & 2.604 \\
$r_p=-0.2$ & 41.5 & \textbf{2.394} & 28.3 & \textbf{2.307} & 35.9 & \textbf{2.221} \\
\hline
%$t_{max}=4$ & 48.3 & 2.831 & 31.7 & 3.235 & 37.2 & 2.908 \\
%$t_{max}=16$ & 52.7 & 32.000 & 35.0 & 32.000 & 41.6 & 32.000 \\
%$t_{max}=16, r_p=0.1$ & 50.4 & 9.471 & 33.6 & 11.238 & 38.4 & 9.717 \\
No handcrafted features & 48.2 & 5.931 & 32.4 & 7.793 & 37.3 & 6.645 \\
DRQN & 48.7 & 3.291 & 33.0 & 6.119 & 40.2 & 5.716 \\
Sequential & 51.4 & 7.834 & 34.1 & 9.318 & 40.8 & 7.423 \\
Video fine-tune & 50.3 & 16.000 & 32.7 & 16.000 & 40.0 & 16.000 \\
\bottomrule
\end{tabular}
\end{center}
\end{small}
\vspace{-2mm}
\caption{Test results for our model based on Alexnet image feature extractor.}
\label{tab:3}
\end{table*}

\subsection{Evaluation Settings}
We evaluate our algorithm with three most commonly used public datasets for multi-shot re-id problem: iLIDS-VID\cite{ilids}, PRID2011\cite{prid} and MARS\cite{mars}.
For iLIDS-VID and PRID2011 dataset, following the setting in \cite{McLaughlinRM16} we randomly split the dataset half-half for training and testing and average the results of 10 runs to make the evaluation stable.
For MARS dataset, we follow the setting by the authors of the dataset. $625$ identities are used for training, and the rest are used for testing. In testing, $1980$ tracklets are preserved for query sets, while the rests are used as gallery sets.

To evaluate performance for each algorithm, we report the Cumulative Matching Characteristic (CMC)  metric. It represents the expectation of the true matching hits in the first top-$n$ ranking. Here we use $n\in\{1, 5, 10, 20\}$ in the evaluations.

\subsection{Ablation Studies}
\label{ref:ablation}

Before comparing our models with previous works, we first conduct ablation studies of some important factors of our method. The results are listed in Table \ref{tab:2} and Table \ref{tab:3} for different settings and datasets. As a baseline, we calculate the averagely pooled features mentioned in Equation \ref{eq:1}. The results of baseline method using all frames are listed in \textbf{All frames} rows.

First let's discuss an important parameter of our model: the reward for unsure action $r_p$. We show the statistics of how many images are used (which is double of the time steps) in each episode in Fig. \ref{fig:episodelength} and corresponding CMC rank $1$ in Table \ref{tab:2} and Table \ref{tab:3}. When $r_p$ is small (negative), the agent will stop early and verify the identities with fewer images. When $r_p$ is big (positive), the agent will be encouraged to be more cautious, requesting more image pairs for better performance. This will help the agent postpone its decision to avoid mistakes caused by imperfect quality like occlusions.
%\footnote{Postponing decisions may cause getting imperfect quality images. However, in one hand, postponing increases the chance of getting perfect images as well as the chance of imperfect images. Assuming that imperfect images are only small amount, such behavior is worthy to have; In the other hand, the agent has history information in $h_t$, this helps it to avoid misjudging. *}
Among all different values of $r_p$, we found that $r_p=0.2$ gives us the most remarkable performance.

We compare the CMC Rank $1$ results of our proposed models with baseline methods in Figure \ref{fig:c1curve}. The dashed green line denotes the \textbf{All frames} setting in Table \ref{tab:2} and Table \ref{tab:3}, while the blue stars denotes the setting that we randomly sample pairs from the tracks, and then averagely pool their features to a track level feature. We vary the number of images sampled to generate the curve.
% And the scatters in different colors are the results generated by our algorithm with different $r_p$, using the same colors as in Figure \ref{fig:episodelength}.
And the yellow squares show the CMC Rank $1$ performance of our model with different values of $r_p$.
We then take a close look of the analysis of the number of images used in these two networks. Not surprisingly, our method uses notably less number of images. Particularly, we can outperform the \textbf{All frames} baselines using only 3\% to 4\% images. We owe the reason to that the average pooling of all the frames may be easily contaminated by some imperfect frames.

In Figure \ref{fig:c1curve}, we also compare the CMC Rank $1$ results of our model with different choices of the maximum time step $t_{max}$. We take three different choices: $t_{max}=4$ (red triangles), $t_{max}=8$ (yellow squares) and $t_{max}=16$ (seafoam blue pentagons) and see how CMC Rank $1$ changes with different values of $r_p$. Comparing among three settings, we find that $t_{max}=8$ gives the best trade-off between number of images used and performance.

%Then we evaluate the choice of network structure for the image feature extractor. We use two different networks: Alexnet\cite{imagenet} and Inception-BN\cite{incptbn} to test our models. In specific, we improve over the \textbf{All frames} baseline of AlexNet remarkably, while we achieve comparable results with state-of-the-art methods using Inception-BN. Compared with Alexnet, Inception-BN provides more accurate image level discriminative features, therefore the agent tends to make a quicker decision. The agent chooses more samples to stop at the end of a episode in AlexNet. %This phenomenon shows the agent can determine whether it has collect enough evidence or not, making us improve more from average pooling baseline compared to Inception-bn.

Next, we compare across different datasets. There are tons of occlusions in iLIDS-VID and MARS datasets. Moreover, there are many mislabeled samples in MARS since the bounding boxes of MARS dataset are machine generated. PRID2011 dataset is much easier compared with the other two datasets. We find that the agent tends to ask for more images in iLIDS-VID and MARS dataset than PRID2011 dataset under the same setting. These two findings coincide with our anticipated behavior of the agent.

\begin{table*}[!ht]
\begin{center}
\begin{small}
\begin{tabular}{l|cccc|cccc|cccc}
\toprule
Dataset & \multicolumn{4}{c|}{PRID2011} &  \multicolumn{4}{c|}{iLIDS-VID} & \multicolumn{4}{c}{MARS}\\
\hline
CMC Rank & 1 & 5 & 10 & 20 & 1 & 5 & 10 & 20 & 1 & 5 & 10 & 20 \\
\hline
RNN-CNN\cite{McLaughlinRM16} & 70 & 90 & 95 & 97 & 58 & \textbf{87} & 91 & 96 &  40 & 64 & 70 & 77\\
ASTPN\cite{abs-1708-02286} & 77 & 95 & \textbf{99} & 99 & \textbf{62} & 86 & \textbf{94} & \textbf{98} & 44 & 70 & 74 & 81\\
Two-Stream\cite{Chung_2017_ICCV} & 78 & 94 & 97 & 99 & 60 & 86 & 93 & 97 & - & - & - & -\\
CNN+XQDA\cite{ZhengSTWWT15} & 77.9 & 93.5 & - & 99.3 & 53.0 & 81.4 & - & 95.1 & 65.3 & 82.0 & - & 89.0 \\
\hline
Alexnet (All frames) & 47.8 & 74.4 & 83.6 & 91.2 & 32.1 & 59.0 & 70.0 & 80.6 & 36.8 & 53.1 & 61.6 & 68.8\\
Alexnet + Ours & 52.6 & 81.3 & 88.4 & 96.3 & 35.1 & 61.3 & 72.1 & 84.0 & 41.2 & 55.6 & 63.1 & 73.3\\
Inception-BN (All frames) & 84.3 & 96.5 & 98.8 & \textbf{99.7} & 60.0 & 85.4 & 92.0 & 96.3 & 68.3 & 83.5 & 88.0 & 90.8\\
Inception-BN + Ours & \textbf{85.2} & \textbf{97.1} & 98.9 & 99.6 & 60.2 & 84.7 & 91.7 & 95.2 & \textbf{71.2} & \textbf{85.7} & \textbf{91.8} & \textbf{94.3}\\
\hline
QAN\cite{LiuYO17} & 90.3 & 98.2 & 99.3 & 100 & 68.0 & 86.8 & 95.4 & 97.4 & - & - & - & -\\
STRN\cite{Zhou_2017_CVPR} & 79.4 & 94.4 & - & 99.3 & 55.2 & 86.5 & - & 97.0 & 70.6 & 90.0 & - & 97.6 \\
\bottomrule
\end{tabular}
\end{small}
\end{center}
\vspace{-2mm}
\caption{Comparisons with other state-of-the-art methods. Please note that the results in last two rows are not directly comparable due to different setting. For more details, please refer to the text.}
\label{tab:comparison}
\end{table*}

Finally there are some more settings worthy trying. We put these experiment results in Table \ref{tab:2} and Table \ref{tab:3} with $r_p=0.2$ and $t_{max}=8$ if not specially mentioned.
\begin{itemize}
	%\item $\mathbf{t_{max}=4}$: Here we show what will happen when $t_{max}$ varies by setting $t_{max}=4$ and keeping other settings the same. We found that there are fewer images used in each episode, but CMC Rank $1$ becomes worse compared with $t_{max}=8$.
    %\item $\mathbf{t_{max}=16}$: Sharing the same motivation with $t_{max}=4$, we change $t_{max}$ for a longer one to $16$. Since the goal of the agent is to get maximum cumulated rewards for each episode, $r_p=0.2$ is too big when $t_{max}=16$ and the agent will always choose ``unsure''. Here we report another result setting $t_{max}=16$ and $r_p=0.1$. The CMC Rank $1$ is close to the result when $r_p=0.2$ and $t_{max}=8$ while using much more images.  \footnote{May report the results of $r=0.1$ to make these results reasonable}
	\item \textbf{No handcrafted features}: We learn the policy without the $3$ dimensions handcrafted distance features, only with image level features and historical information. CMC Rank $1$ drops a lot and the agent will tend to make a quicker choice.
    \item \textbf{DRQN}: We try to replace the last fc layer with a LSTM layer as in \cite{DRQN} to gather historical features instead of the method we described in \ref{sec:singlereid}. The results are worse compared with our proposed method.
    \item \textbf{Sequential}: Instead of feeding the agent with random ordered images, we try to provide the images sequentially started from the beginning of the sequences. The results are worse compared with random order.
    \item \textbf{Video fine-tune}: Here we randomly sample $8$ images from each sequence, averagely pool the features and use this sequence level feature to fine-tune the CNN as described in Sec.~\ref{sec:singlereid}. This model gets a slightly worse CMC Rank $1$ performance, but uses more images.
\end{itemize}

%Here two different strategies are applied for baseline, the first one is to use all the images from each identity(shown as the blue horizontal line in Figure \ref{fig:c1curve}), the second one is to sample equal amount of image for each identity(shown as the green triangles in Figure \ref{fig:c1curve}). We enumerate $r_p$ for each dataset and image feature extractor and plot the results with red circles. Accurate results can be found in Table\ref{tab:2} and Table \ref{tab:3}. In Figure \ref{fig:episodelength}, we show how many images are used for each episode with different choice of $r_p$ for different datasets and image feature extractors.

\begin{figure*}[!ht]
\begin{center}
   \includegraphics[width=0.33\linewidth]{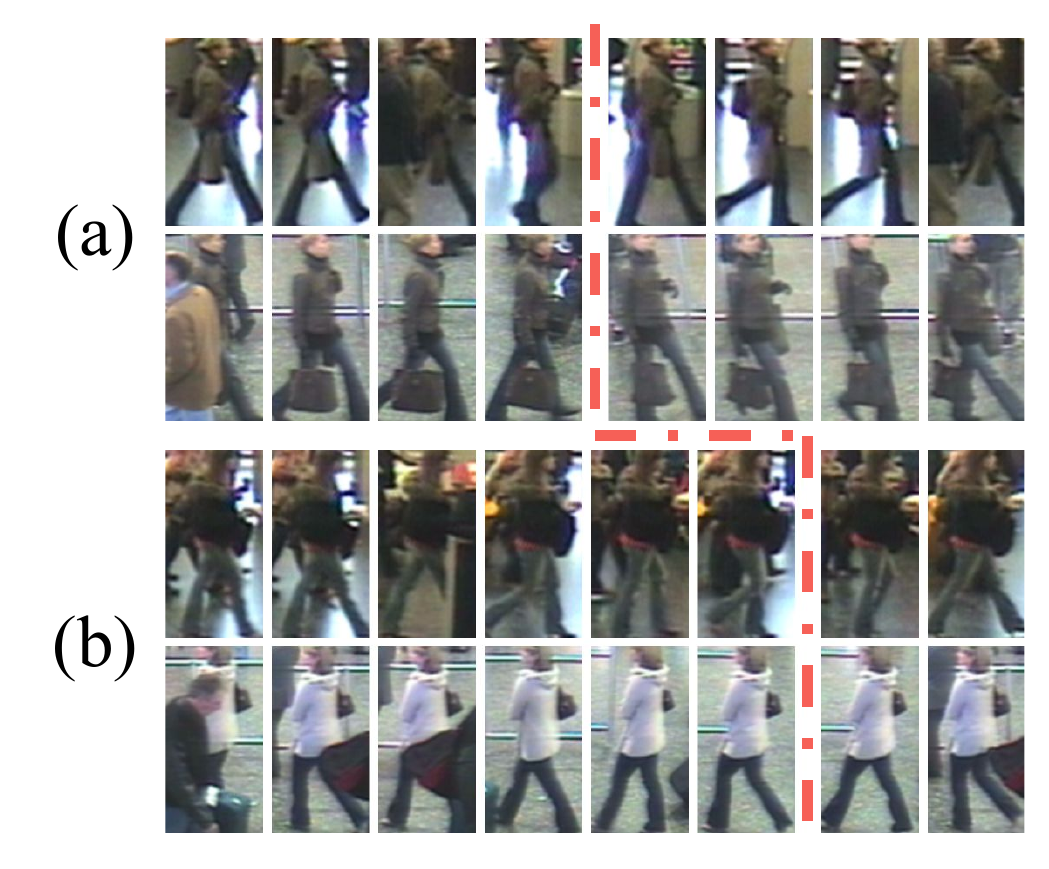}
   \begin{minipage}[b]{0.64\linewidth}
    \begin{center}
     %\subfloat[]{\includegraphics[width=0.49\linewidth]{figure/eg_a.pdf}}
     \subfloat[]{\includegraphics[width=0.45\linewidth]{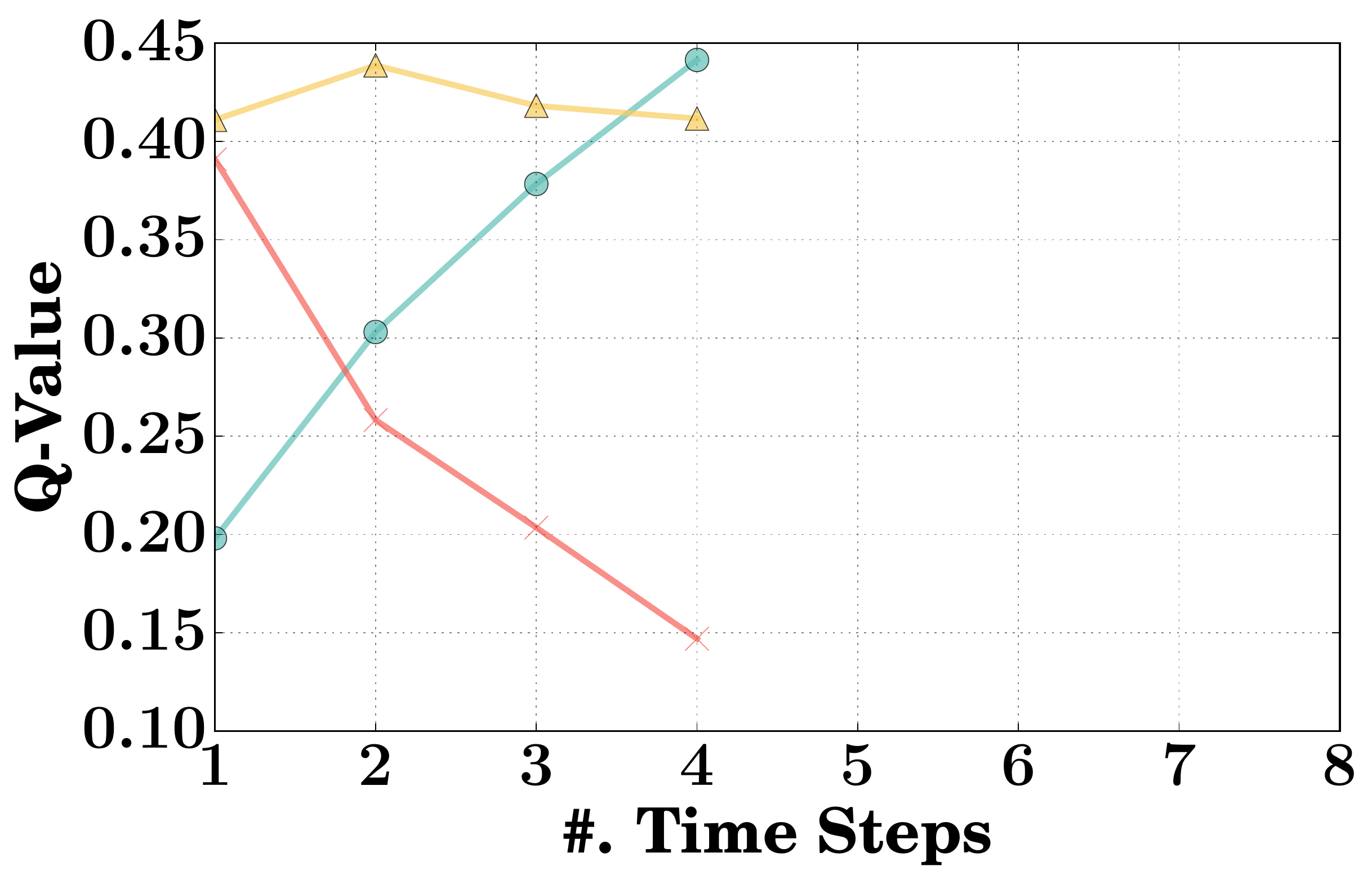}}
     \subfloat[]{\includegraphics[width=0.45\linewidth]{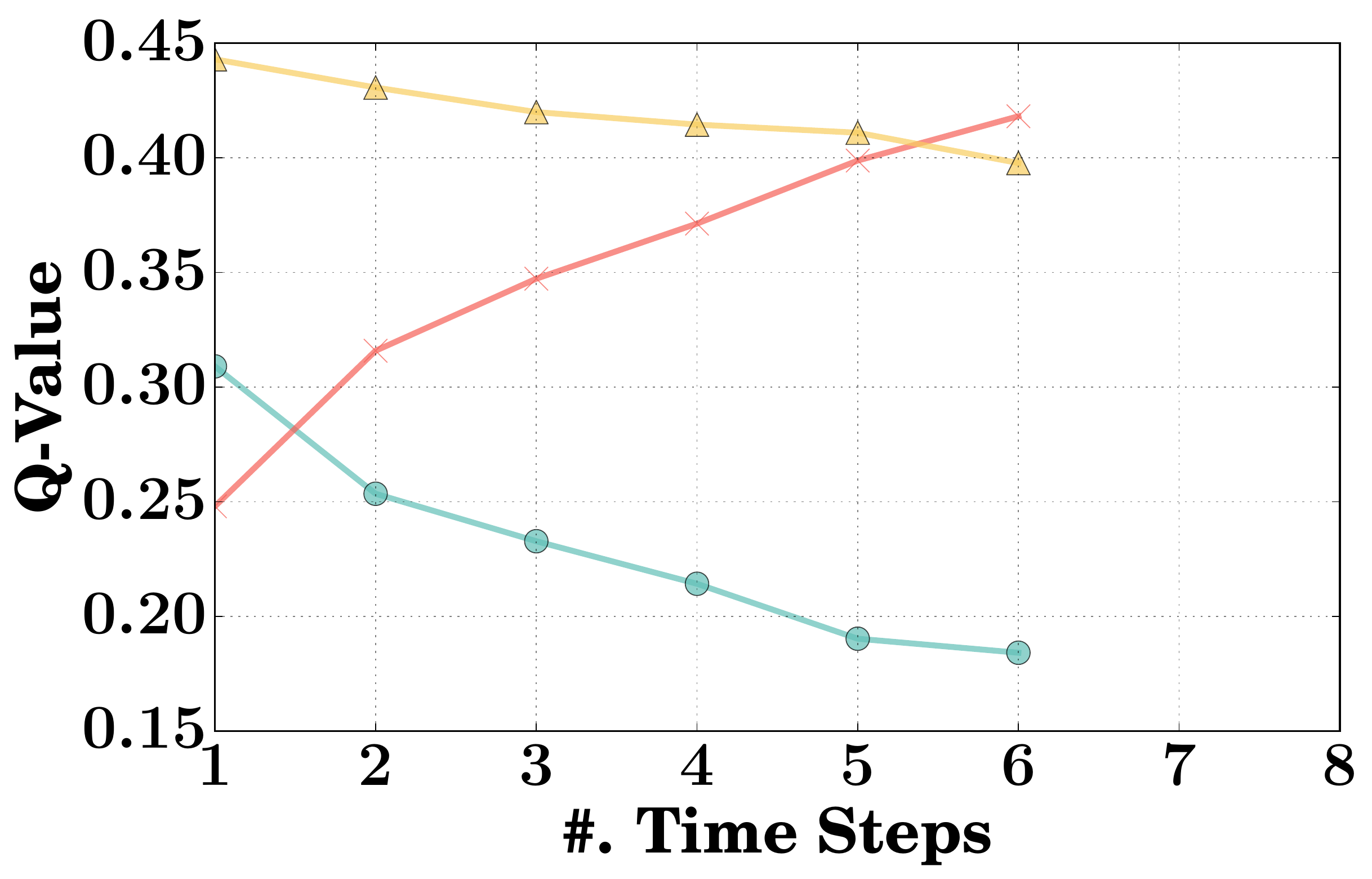}}\hfill
     %\subfloat[]{\includegraphics[width=0.49\linewidth]{figure/eg_d.pdf}}\hfill
     \includegraphics[width=0.6\linewidth]{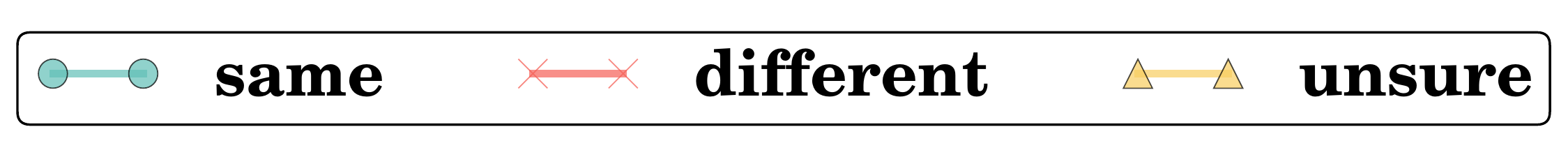}
    \end{center}
   \end{minipage}
\end{center}
	\vspace{-4mm}
   \caption{Some example episodes generated by our model. All the sampled images for each identity are listed on the left with a red dashed line splits used images and unused images. On the right side, normalized Q values for each example are shown.}
\label{fig:qualitative}
	\vspace{-1mm}
\end{figure*}

\subsection{Comparisons with State-of-the-art Methods}
Table \ref{tab:comparison} summarizes the CMC results of our model and other state-of-the-art multi-shot re-id methods. Here we use the setting of $r_p=0.2$ since this setting is the most accurate according to the evaluations in previous section. CNN-RNN\cite{McLaughlinRM16}, ASTPN\cite{abs-1708-02286}, STRN\cite{Zhou_2017_CVPR} and Two-Stream\cite{Chung_2017_ICCV} are four different methods based on RNN time series model and more advanced attention mechanism. CNN-XQDA\cite{ZhengSTWWT15} and QAN\cite{LiuYO17} train discriminative embeddings of images and apply different pooling methods. Among them, CNN-RNN\cite{McLaughlinRM16}, ASTPN\cite{abs-1708-02286} and Two-Stream\cite{Chung_2017_ICCV} use both image and explicit motion features (optical flow) as inputs for deep neural network.

Here QAN\cite{LiuYO17} uses their own extra data for training. STRN\cite{Zhou_2017_CVPR} uses MARS pre-trained model to train PRID2011 and iLIDS-VID. Therefore, their methods cannot be fairly compared with other methods. We just list their results for reference.

For PRID2011 dataset, our method outperforms all other methods, improves the CMC Rank 1 about $5\%$ compared with best state-of-the-art methods. For iLIDS-VID and MARS dataset, our results are at least comparable or even better than the compared methods.
%For iLIDS-VID and MARS dataset, the occurrence of more occlusions makes the image feature extractor perform poorly. The baseline CMC rank $1$ is about $24\%$ lower for iLIDS-VID dataset and $16\%$ lower for MARS dataset compared with PRID2011 dataset with Inception BN image feature extractor. Since our model is based on image feature extractors, our improvement on iLIDS-VID and MARS from baseline is not that much. For iLIDS-VID dataset, our model beats all the state-of-the-art methods except for ASTPN\cite{abs-1708-02286}. ASTPN\cite{abs-1708-02286} uses spatial pooling to pay more attention to specific region. This is very helpful to deal with occlusions. For MARS dataset, we improved the state-of-the-art about $0.6\%$ for CMC Rank $1$ compared with all the other methods. This shows our model is robust enough to generalize to huge datasets.
For CMC Rank $5$, $10$ and $20$, the trends are similar to Rank $1$.

Note that all the other methods use all images for verification. \emph{Our proposed model uses only 3\% to 6\% images for each track pairs on average to obtain these encouraging performance.}

\subsection{Qualitative Results}
In Figure \ref{fig:qualitative}, two representative episodes are shown. We can see the change of the Q values for the agent in dynamic environment. Softmax function is applied to normalize the Q values. (a) shows an example episode with the same person, while (b) shows one with different persons. These two episodes end with different length. Severe occlusions happen in the early pairs of (a) and (b). After the occlusions disappear, the agent gradually collects information and corrects its decisions. After fed with several image pairs of better quality, the agent is confident enough to make the correct choices eventually.
%The agent stops right after the occlusions disappear. For (d), the these two persons have similar appearances but they are different persons, the agent had a high but not enough confidence to choose {\it different}. Then another person appears in the second and third pair makes the agent hesitate and reduce its confidence for the {\it different} action. After receiving more images, the agent has collected enough information and choose {\it different} eventually.

%-------------------------------------------------------------------------
\section{Conclusion}
In this paper we have introduced a novel approach for multi-shot pedestrian re-identification problem by casting it as a pair by pair decision making process. Thanks to reinforcement learning, we could train an agent for such task. Specifically, it receives image pairs sequentially, and output one of the three actions: {\it same, different} or {\it unsure}. By early stop or decision postponing, the agent could adjust the budget needs to make confident decision according to the difficulties of the tracks.

We have tested our method on three different multi-shot pedestrian re-id datasets. Experimental results have shown our model can yield competitive or even better results with state-of-the-art methods using only $3\%$ to $6\%$ of images. Furthermore, the Q values outputted by the agent is a good indicator of the difficulty of image pairs, which makes our decision process is more interpretable.

Currently, the weight for each frame is determined by the Q value heuristically, which means the weight is not guided fully by the final objective function. More advanced mechanism such as attention can be easily incorporated into our framework. We leave this as our future work.
%The Q values generated by Q function may contain rich information for image quality and context information. Our future work lies in applying these Q values to learn attention and gather temporal information from videos. 

\section{Acknowledgement}
The work was supported in part by the National Basic Research Program of China (Grant No. 2015CB856004), the Key Basic Research Program of Shanghai Municipality, China (15JC1400103,16JC1402800)

{\small
\bibliographystyle{ieee}
\bibliography{refs}
}

\end{document}